\documentclass{article}

\PassOptionsToPackage{numbers,compress}{natbib}
\usepackage[eandd, final]{neurips_2026}

\usepackage[utf8]{inputenc} 
\usepackage[T1]{fontenc}    
\usepackage{hyperref}       
\usepackage[table]{xcolor}
\usepackage{fontawesome5}

\definecolor{linkblue}{RGB}{0,90,180}
\hypersetup{colorlinks=true, urlcolor=linkblue, linkcolor=black, citecolor=black}
\usepackage{caption}
\usepackage{url}            
\usepackage{booktabs}       
\usepackage{amsfonts}       
\usepackage{microtype}      
\usepackage{amssymb}
\usepackage{adjustbox}
\usepackage{multirow}
\usepackage{graphicx}
\makeatletter
\@ifundefined{Gread@transgroupfalse}{}{\let\Gread@transgrouptrue\Gread@transgroupfalse}
\makeatother
\usepackage{enumitem}
\usepackage{nicematrix}
\usepackage[accsupp]{axessibility}  
\newcommand{\mypar}[1]{\vspace{0mm}\noindent\textbf{#1}}

\newcommand{\symbolimg}[2][0.3cm]{%
  \ensuremath{\vcenter{\hbox{\includegraphics[height=#1]{#2}}}}%
}
\newcommand{\gemini}{\symbolimg[0.3cm]{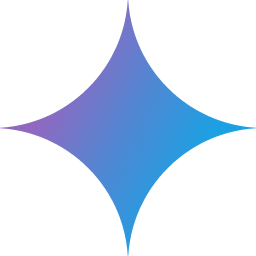}}
\newcommand{\claude}{\symbolimg[0.3cm]{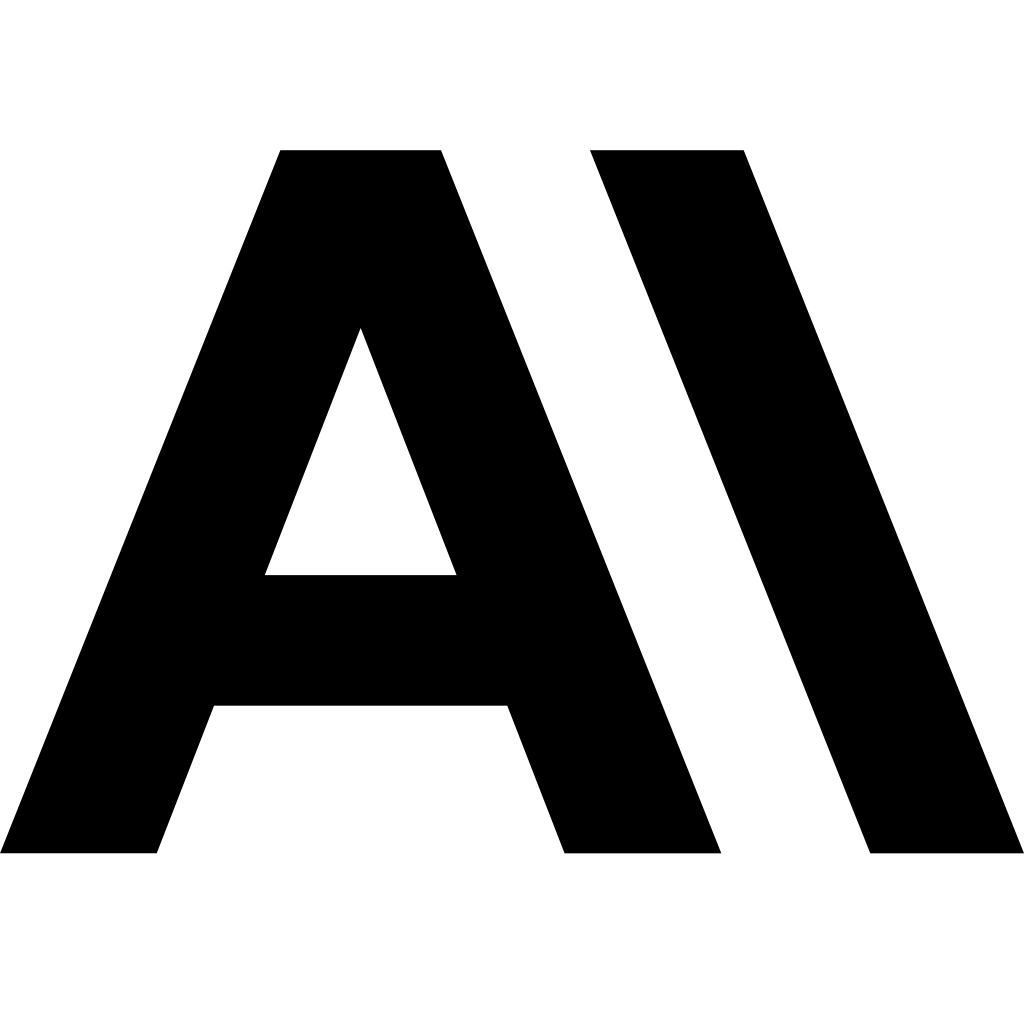}}
\newcommand{\openai}{\symbolimg[0.3cm]{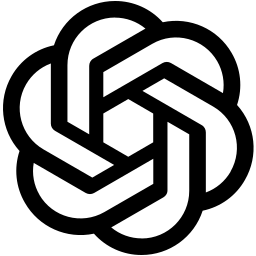}}
\newcommand{\qwen}{\symbolimg[0.3cm]{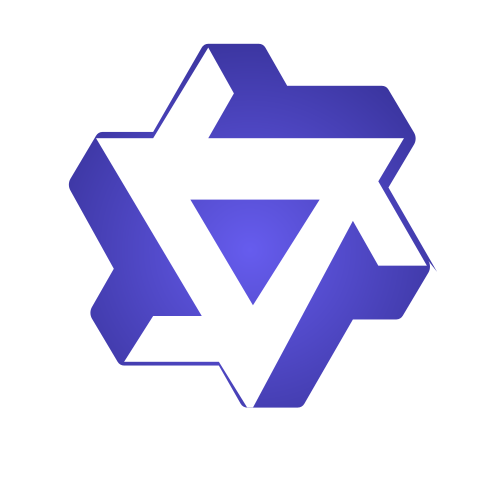}}
\newcommand{\llava}{\symbolimg[0.3cm]{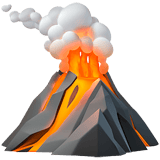}}
\newcommand{\gemma}{\symbolimg[0.3cm]{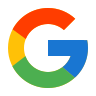}}

\usepackage{transparent}
\newcommand{\faded}[1]{\texttransparent{0.5}{#1}}

\usepackage{xspace}
\newcommand{\benchmark}{\textsc{EditJudge-Bench}\xspace}

\usepackage[most]{tcolorbox}

\newtcolorbox{promptbox}[1]{
  colback=gray!4,
  colframe=gray!35,
  title=#1,
  fonttitle=\bfseries,
  boxrule=0.4pt,
  arc=2pt,
  left=6pt,
  right=6pt,
  top=5pt,
  bottom=5pt,
  before skip=6pt,
  after skip=8pt
}

\usepackage{tikz}

\definecolor{navy}{HTML}{1F3A5F}
\colorlet{genGray}{navy!5!white}            
\colorlet{probGray}{navy!15!white}          
\colorlet{supGray}{navy!20!white}           
\colorlet{propGray}{navy!10!white}          

\newcommand{\tbgap}{\hspace{0.25em}}

\newcommand{\tbparadigmType}[2]{\multicolumn{7}{l}{\rowcolor{#1}\textit{#2}\rule[-1.3ex]{0pt}{3.2ex}}}
\newcommand{\tbparadigmC}[2]{\multicolumn{12}{l}{\rowcolor{#1}\textit{#2}\rule[-1.3ex]{0pt}{3.2ex}}}
\newcommand{\tbparadigmCorr}[2]{\multicolumn{5}{l}{\rowcolor{#1}\textit{#2}\rule[-1.3ex]{0pt}{3.2ex}}}

\newcommand{\rGen}{\rowcolor{genGray}}
\newcommand{\rProb}{\rowcolor{probGray}}
\newcommand{\rSup}{\rowcolor{supGray}}
\newcommand{\rProp}{\RowStyle[color=black!70, rowcolor=propGray]{}}

\NiceMatrixOptions{
  custom-line = {ccommand = tbdotcline, tikz = {densely dotted, gray!60}, total-width = 0pt},
  custom-line = {command = tbdashline, tikz = {dashed, gray!60},         total-width = 5pt},
  custom-line = {letter = I, tikz = {gray!70, line width = 0.6pt},        total-width = 0pt}
}
\newcommand{\tbdots}[1]{\tbdotcline{2-12}}   
\newcommand{\tbsepc}[1]{\tbdashline}         

\newcommand{\projectlink}{%
  \href{https://jenspars.github.io/EditJudge-Bench/}{%
    {\color{navy}\faGlobe}\hspace{0.35em}\texttt{jenspars.github.io/EditJudge-Bench/}}}

\ExplSyntaxOn
\msg_if_exist:nnTF { nicematrix } { key~color-inside }
  { \NewDocumentCommand \NiceTab { m } { \begin{NiceTabular}{#1} } }
  { \NewDocumentCommand \NiceTab { m } { \begin{NiceTabular}{#1}[colortbl-like] } }
\ExplSyntaxOff

\title{\texorpdfstring{\raisebox{-0.25em}{\includegraphics[height=1.3em]{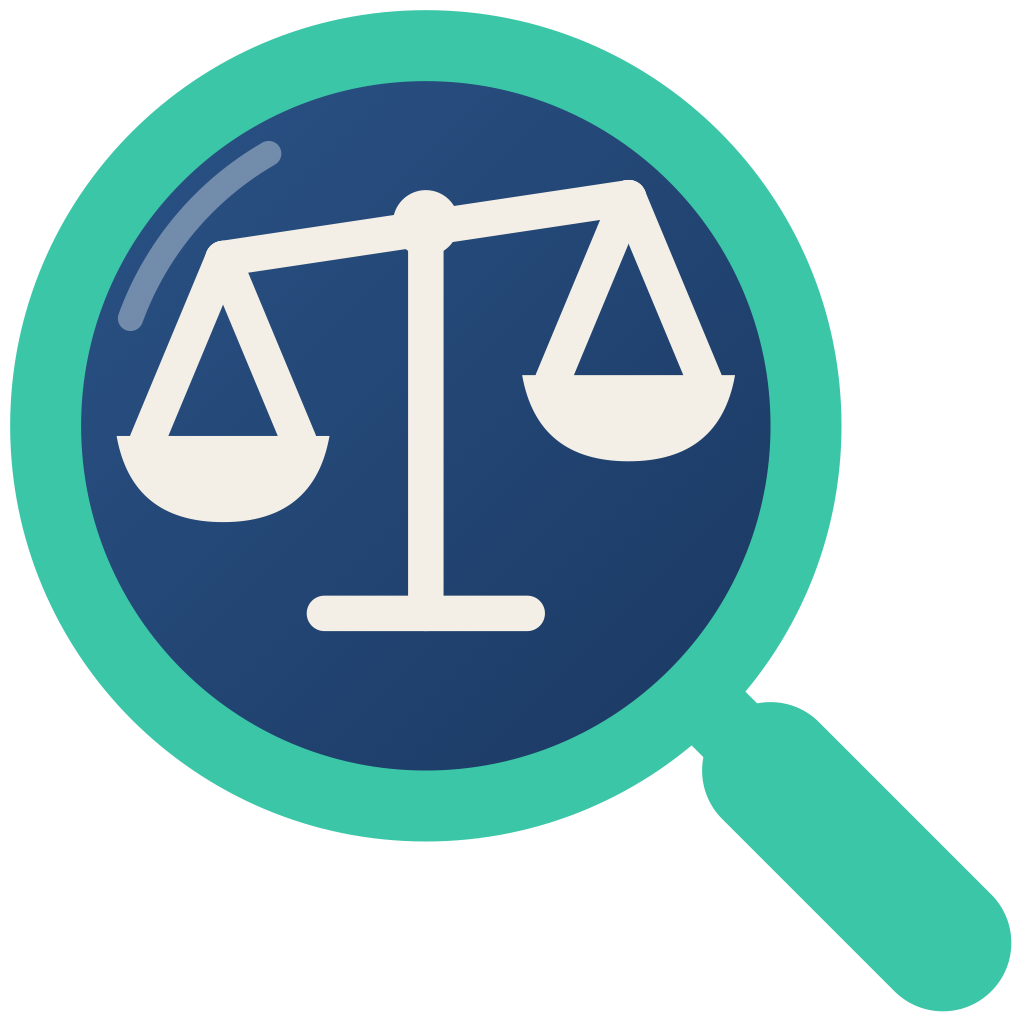}}\hspace{0.3em}}{}EditJudge-Bench: Auditing VLM Image-Edit Judges with Synthetic Ground Truth}

\author{
  Jens Parslov$^{1,2}$ \qquad
  Marco Schouten$^{1,2}$ \qquad
  Dim~P.~Papadopoulos$^{1,2}$ \\
  $^{1}$Technical University of Denmark \qquad
  $^{2}$Pioneer Centre for AI \\
  {\small\texttt{\{jenpa, marscho, dimp\}@dtu.dk}} \\[0.3em]
  {\projectlink}
}

\begin{document}

\maketitle

\vspace{-2.8em}
\begin{center}
  \includegraphics[width=\textwidth]{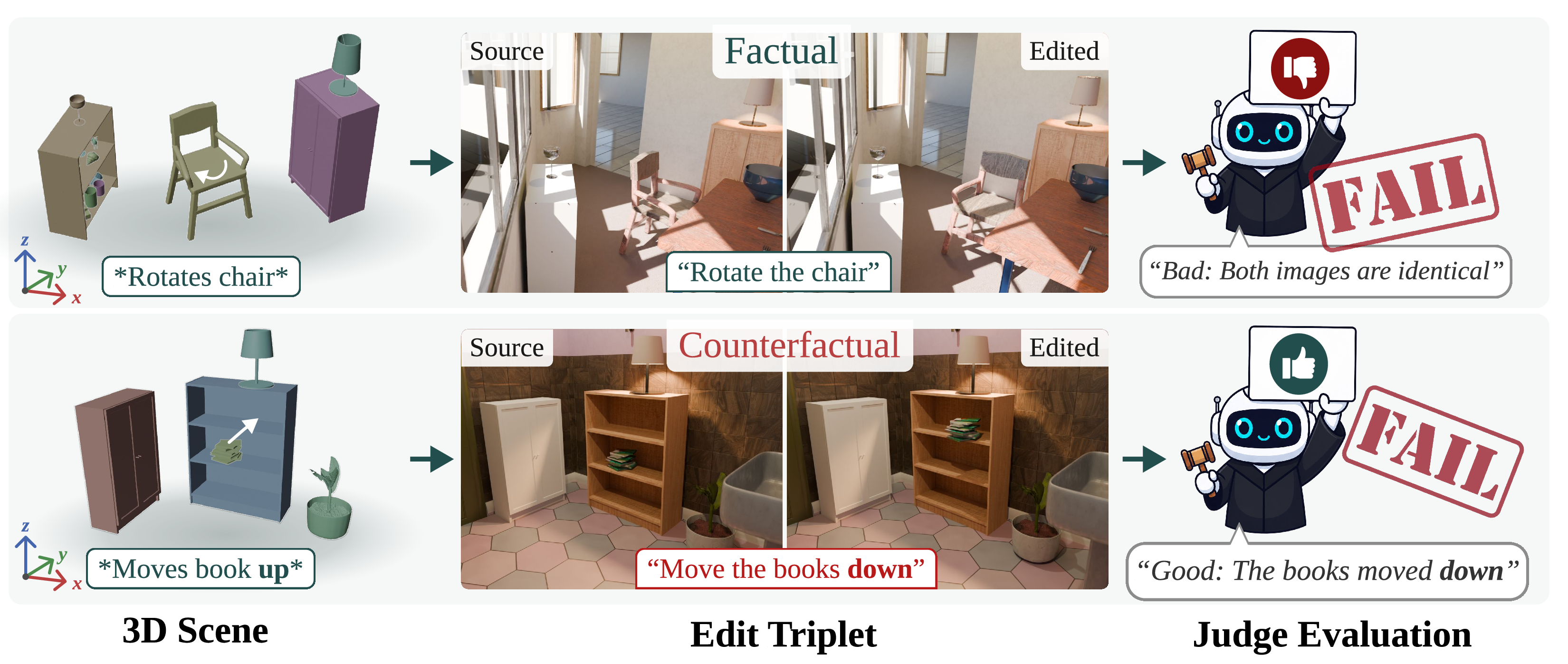}
  \vspace{-1.2em}
  \captionof{figure}{\textbf{Auditing VLM image-edit judges with \benchmark.}
    We procedurally alter 3D scenes (left) producing (source, edited, instruction) triplets (middle). We construct factual (top) and counterfactual (bottom) triplets to
    audit VLM image-edit judges (right), exposing reliability gaps in current VLM-as-a-judge evaluation protocols.}
  \label{fig:teaser}
\end{center}
\vspace{-0.8em}

\begin{abstract}
Vision-language models (VLMs) are increasingly used as automated judges for image-editing benchmarks, yet their reliability is poorly understood. Auditing these judges is difficult because real edited images lack controlled ground truth and entangle the requested changes with model artefacts, unintended side effects, and ambiguous degrees of success. In this paper, we introduce EditJudge-Bench, a diagnostic benchmark of 1,382 synthetic before/after image pairs rendered in Blender, with edits parameterised across nine categories derived from real user requests and exact ground truth from known 3D scene parameters. Because every edit is generated by a known intervention on a controlled scene, EditJudge-Bench evaluates VLM judges as binary verifiers of whether an instruction was correctly executed, isolating judge perception from editor quality. Across 25 judge configurations, we find substantial reliability gaps: (a) the widely adopted VIEScore template reaches only 69.0\% AUROC with GPT-4o; (b) fine-tuned judges score 51.7–66.3\%; spatial edits such as rotation and shape remain challenging; (c) prompt-formatting choices like image–text ordering shift performance by up to 14.1 points. 
Despite using only synthetic data, judge rankings on EditJudge-Bench correlate strongly with two real-image human-preference benchmarks ($\rho$=0.95 and $\rho$=0.71), indicating that controlled audits surface failure modes relevant to real evaluation.
\end{abstract}

\section{Introduction}
\label{sec:introduction}

Instruction-based image editing systems can now modify scenes, swap objects, and adjust visual attributes from natural-language prompts~\cite{brooks2023instructpix2pix, sheynin2024emuedit, zhang2024magicbrush}. As these systems become more capable, reliable evaluation becomes increasingly important. Human evaluation remains the gold standard, but it is expensive, slow, and difficult to reproduce. Consequently, many benchmarks use Vision-Language Models (VLMs) as automated judges~\cite{ku2024viescore, liu2025step1xedit, ye2025imgedit, qian2025giebench, huang2024i2ebench}, making VLM scores part of the default evaluations.

This reliance raises a fundamental question: can these judges reliably verify the edits they are asked to score? Existing validation protocols typically measure correlation with human preferences over a fixed pool of editing-model outputs. Such correlation is useful, but indirect. A judge may agree with humans on average by detecting broad image quality, salient artifacts, or obvious semantic changes, while still failing to perceive precise edits such as rotation, shape change, subtle attribute shift, or whether a preservation constraint was satisfied. 

Examining this failure mode is challenging because real editing-model outputs lack controlled ground truth. The requested edit is entangled with model-specific artefacts, unintended modifications, and ambiguous degrees of success. If a judge assigns a poor score, it is often unclear whether the judge failed, the edit request was ambiguous, or the edited image was genuinely flawed. This ambiguity is especially problematic because edit evaluation requires visual differencing across two images, grounding the observed change in the instruction, and detecting unintended changes to preserved content.  VLMs already struggle with spatial reasoning and fine-grained visual discrimination in single images~\cite{tong2024eyes}; therefore, before/after edit evaluation may expose even sharper limitations.

To isolate these factors, we need diagnostic benchmarks in which the intended edit, its magnitude, and the preserved content are known exactly. We therefore introduce \benchmark, a controlled benchmark for auditing VLM-based image-editing judges. \benchmark contains 1,382 synthetic before/after image pairs rendered in Blender, with parameterised edits spanning nine edit types and exact ground truth from 3D scene parameters. Because \benchmark audits judges rather than ranking editors, it uses controlled synthetic edits and binary ground-truth correctness labels. These labels are not intended to replace holistic human judgments of edit quality; instead, they provide targeted probes of whether a judge can distinguish instruction-consistent from instruction-inconsistent changes. For example, given the instruction ``rotate the chair to the left,'' a factual pair contains the requested rotation, while a counterfactual pair changes an irrelevant property. This construction separates judge failures from the ambiguities of real editing-model outputs.

Our experiments reveal substantial reliability gaps in current evaluation methods. The widely used VIEScore template~\cite{ku2024viescore} achieves only 69.0\% AUROC with GPT-4o, while fine-tuned judge models range from 51.7\% to 66.3\% AUROC. Prompt-formatting choices can shift performance by up to 14.1 points, and several judges exhibit systematic blind spots for spatial edits, and tend to miss small or low-salience changes when verifying instructions. Surprisingly, a simple log-probability baseline outperforms several complex zero-shot and fine-tuned judge protocols. To validate that audits on synthetic data are informative for real-image evaluation, we further show that judge rankings on \benchmark correlate strongly ($\rho$=0.95) with rankings on GenAI-Bench~\cite{jiang2024genai}, a human-preference benchmark over real edited images.

\noindent Our contributions are:

\begin{enumerate}
    \item We introduce \benchmark, a diagnostic benchmark of 1,382 controlled before/after image pairs rendered in Blender, with exact ground truth from 3D scene parameters and an edit taxonomy derived from real user requests.

    \item We present a procedural generation pipeline for constructing factual and counterfactual edit triplets whose correctness is known by construction, enabling controlled audits of VLM-based image-editing judges.

    \item We show that widely used VLM judges are substantially less reliable than their adoption suggests: VIEScore with GPT-4o reaches only 69.0\% AUROC, fine-tuned judges range from 51.7\% to 66.3\%, and prompt-formatting choices can shift performance by up to 14.1 points.

    \item We identify systematic judge failure modes, including weak visual differencing, spatial blind spots, and salience bias on subtle edits, and find that a simple log-probability baseline outperforms several more complex zero-shot and fine-tuned judge protocols.
\end{enumerate}

    \section{Related work}
    \label{sec:relwork}
  
\noindent\textbf{Image editing models.}
Early approaches operated on paired training data~\cite{isola2017image} or required per-image optimization~\cite{kawar2023imagic}. Instruction-based editing removed the need for manual masks or paired examples~\cite{brooks2023instructpix2pix, sheynin2024emuedit, zhang2024magicbrush}, enabling users to specify edits in natural language. Recent systems extend this to an increasingly diverse set of edit types, including spatial rearrangement, appearance and style transfer, geometric transformation, lighting adjustment, and compositional changes~\cite{zhao2024ultraedit, yu2025anyedit}. As the space of supported edits has grown, so has the need for evaluation methods that can assess whether each edit was performed correctly. This growing diversity makes aggregate evaluation scores increasingly difficult to interpret: a metric that is reliable for appearance edits may still fail on spatial, geometric, or magnitude-sensitive edits.

\noindent\textbf{Evaluation of image editing.}
Some benchmarks compare model outputs against ground-truth references obtained from editing models~\cite{zhang2024magicbrush, yu2025anyedit} or human annotators~\cite{sushko2025realedit}, but image editing is a one-to-many task where a single reference is inherently restrictive, and the references themselves inherit the biases of their source. Model-generated references share the failure modes of the generators, while human references vary in quality and lack diversity across edit types. Reference-free CLIP-based metrics~\cite{radford2021learning} avoid this issue~\cite{brooks2023instructpix2pix, sheynin2024emuedit, zhao2024ultraedit, brack2024ledits}. However, their global embeddings behave like bag-of-words representations and perform poorly on spatial and compositional reasoning~\cite{koishigarina2025clip, yuksekgonul2022and}, making them unsuited for localised edit assessment. These limitations motivated the use of Vision-Language Models (VLMs) as judges. VIEScore~\cite{ku2024viescore} introduces prompt templates for semantic consistency and perceptual quality, and recent state-of-the-art benchmarks routinely evaluate edits with VLM-as-a-judge~\cite{wu2025krisbench, liu2025step1xedit, sushko2025realedit, luo2025editscore, wan2025motionedit}. Some approaches fine-tune smaller models to replicate larger ones~\cite{ye2025imgedit, luo2025editscore}, while others train dedicated reward models on human preferences~\cite{qu2025ie, wu2025editreward, luo2025editscore} or combine VLMs with specialist perceptual models~\cite{wang2025i2i, sun2025iebench, qian2025giebench}. Benchmarks have also grown more fine-grained, moving from uniform metrics across all edit types~\cite{zhang2024magicbrush, zhao2024ultraedit} to specialised scoring dimensions~\cite{sun2025iebench, wang2025i2i} and targeted evaluation of specific capabilities such as motion alignment~\cite{wan2025motionedit}, knowledge plausibility~\cite{wu2025krisbench}, and spatial reasoning~\cite{zhao2025envisioning}. These works improve how image editing models are evaluated, but they generally do not isolate whether the VLM judge itself can perceive each instructed edit under controlled ground truth. Our work is complementary: rather than proposing another metric for ranking image editors, we evaluate whether VLM judges can reliably verify controlled edits with known ground truth.

\noindent\textbf{Limitations of VLM judges.}
Using language models as judges is well-established for text evaluation~\cite{zheng2023judging, dubois2023alpacafarm, gu2024survey}, but extending it to vision introduces challenges that text does not face. VLMs exhibit systematic perceptual shortcomings. They inherit visual encoder blind spots that cause failures on basic patterns such as orientation, counting, and spatial relationships~\cite{tong2024eyes, fu2024blink, rahmanzadehgervi2024vision}, and they hallucinate object properties and spatial configurations even when the visual evidence is unambiguous~\cite{li2023evaluating, guan2024hallusionbench}. Existing validation of VLM editing judges typically compares judge scores with human preferences over outputs from a fixed set of editing models~\cite{jiang2024genai, wu2025editreward, luo2025editscore}. This tests aggregate agreement on a particular output distribution, but it does not test perceptual correctness. A judge can correlate with humans while failing to identify whether the requested object moved, changed shape, or preserved irrelevant content. This limitation is amplified when the same small set of editing models defines the preference pool. If those models share artefacts or failure modes, a judge may learn cues associated with model quality rather than the instructed visual change itself. Even dedicated reward models trained on hundreds of thousands of preference pairs~\cite{wu2025editreward} share this limitation. The same editing models appear in both training and evaluation, so generalisation beyond those distributions remains untested. A further concern is reproducibility: many benchmarks rely on proprietary judges whose model snapshots are updated or deprecated over time, making reported scores difficult to regenerate~\cite{chen2023chatgpt, pozzobon2023challenges}. For text-only LLM judges, diagnostic benchmarks with objective correctness labels already exist~\cite{tan2024judgebench, zeng2023evaluating}. Benchmarks such as MLLM-as-a-Judge~\cite{pmlr-v235-chen24h}, MJ-Bench~\cite{chen2025mj}, and Multimodal RewardBench~\cite{yasunaga2025multimodal} evaluate VLM judges on general vision-language tasks and text-to-image generation, but do not target before/after edit verification. For VLM edit judges, EditInspector~\cite{yosef2025editinspector} relies on crowdsourced labels over diffusion-edited images, entangling instruction verification with synthesis artefacts and annotator subjectivity. \benchmark instead derives ground truth from the underlying 3D scene intervention rather than post-hoc human judgement.

\noindent\textbf{Synthetic datasets} are a standard tool for evaluating vision systems where real-world ground truth is impractical to obtain. For judge auditing, this control is especially valuable because it separates evaluator errors from ambiguity or artefacts in the edited image. Rendered scenes with known properties enable diagnostic evaluation with controlled ground truth, as demonstrated across visual reasoning~\cite{johnson2017clevr}, optical flow~\cite{butler2012naturalistic, mayer2016large}, long-term point tracking~\cite{zheng2023pointodyssey}, and inverse rendering~\cite{li2021openrooms}. This principle has also been applied to image difference captioning~\cite {park2019robust, liu2025omnidiff, kim2026vlmsubtlebench}. CLEVR-Change~\cite{park2019robust} and OmniDiff~\cite{liu2025omnidiff} generate synthetic before/after pairs to benchmark models that describe visual changes. However, these datasets target \emph{describing} what has changed and not \emph{scoring} whether an instructed edit was correctly performed. Moreover, their scene complexity and change taxonomies do not align with the edit categories of modern models. Our benchmark fills this gap by using Blender-rendered image pairs with exact ground-truth changes to evaluate VLM judges on specific edit types expected of modern models. OBJect~\cite{michel2023object} is most similar in design, rendering procedural Blender scenes for translation, rotation, insertion, and removal edits, though it trains and evaluates editing models rather than auditing judges.

\section{Procedural generation for editing pairs}
\label{sec:pipeline}

\begin{figure}[t]
    \centering
    \makebox[\textwidth]{\includegraphics[width=1\textwidth]{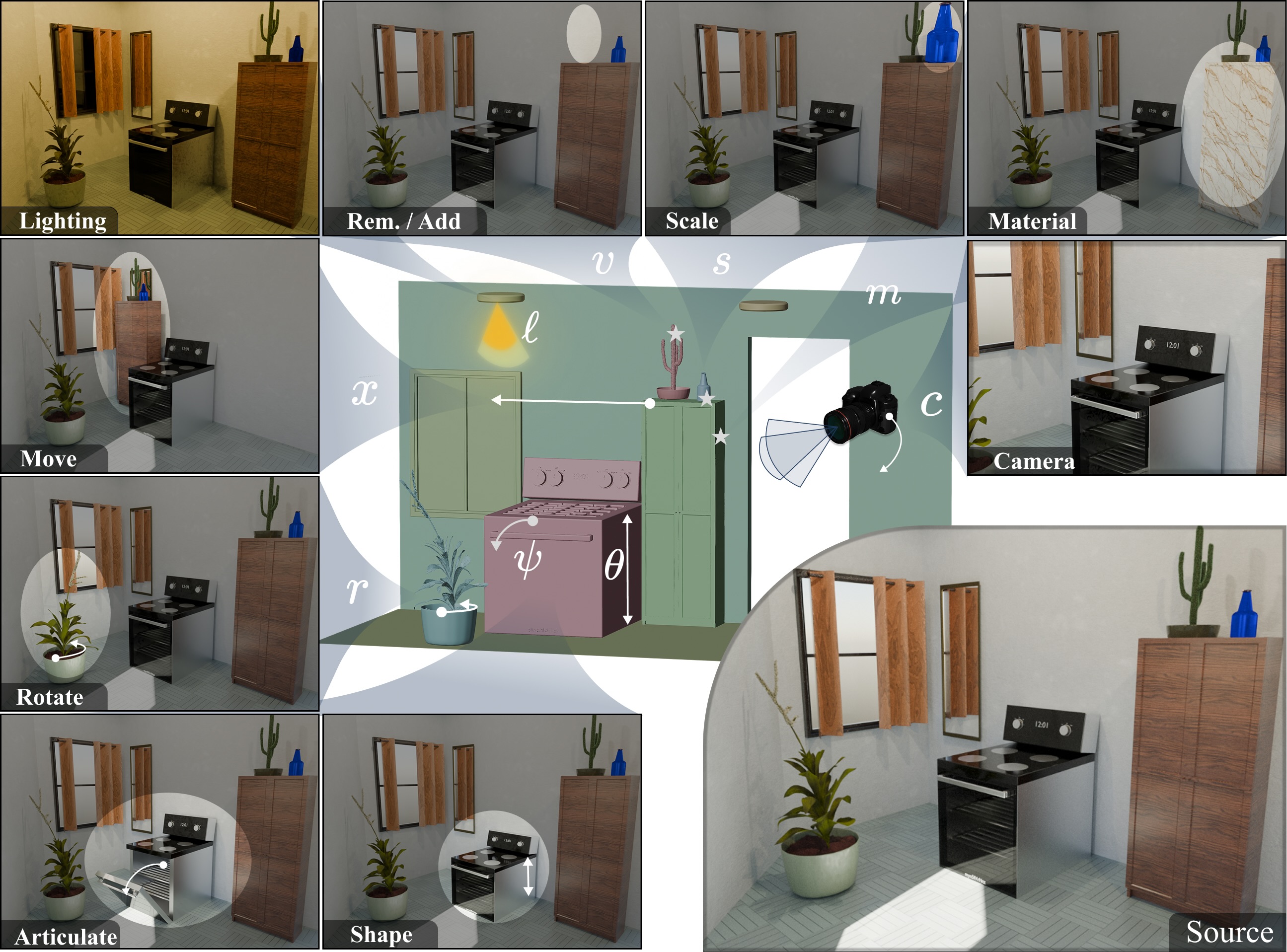}}
    \caption{ \textbf{Overview of the nine edit types in \benchmark.}
    The central inset shows the underlying Blender scene parameter modified by each edit: object position $\mathbf{x}$, rotation $\mathbf{r}$, scale $\mathbf{s}$, shape parameter $\mathbf{\theta}$, articulation joint $\mathbf{\psi}$, material $\mathbf{m}$, visibility $\mathbf{v}$, lighting state $\mathbf{l}$, and camera state $\mathbf{c}$. Surrounding images show one rendered edit from the source image per category. Because each edit is a deterministic parameter change, magnitude and direction are known exactly.
    }
\label{fig:pipeline} 
\end{figure}

\begin{figure}[t]
    \centering
    \includegraphics[width=\textwidth]{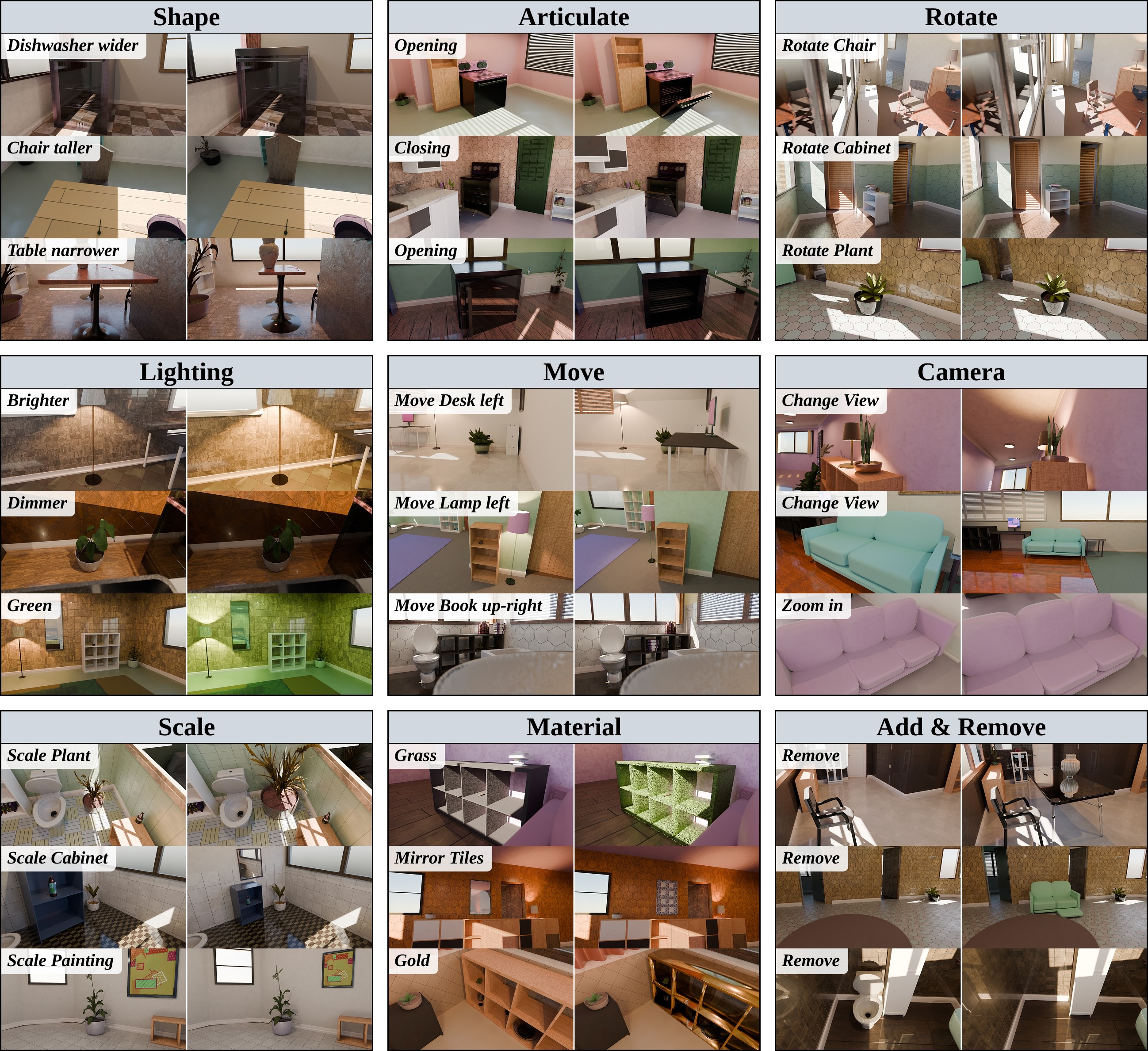}
    \caption{\textbf{ Example image pairs in \benchmark.} For each edit type, we show three original/edited image pairs from our benchmark.}
    \label{fig:examples}
\end{figure}

Auditing a judge requires ground truth that is independent of the systems being judged. When an editing model generates an output, the result entangles the instructed change with model-specific artifacts and hallucinations, imprecise geometry, or unintended side effects.
This makes it impossible to isolate what the judge actually perceives. 
We sidestep this by generating \emph{before/after} pairs entirely in Blender~\cite{Blender2026}, where every edit is a deterministic parameter change in a known 3D scene. Unlike existing editing pipelines, this allows us to guarantee that (i) the edit is \emph{exactly} what the instruction describes, and (ii) everything outside the edit is identical, eliminating confounds. We first derive an edit taxonomy grounded in real user requests (Sec.~\ref{sec:taxonomy}), then assemble diverse 3D scene assets (Sec.~\ref{sec:assets}), and generate edits procedurally with rejection sampling (Sec.~\ref{sec:edit_generation}). Finally, we construct the final benchmark with factual and counterfactual instructions.

\subsection{Edit taxonomy}
\label{sec:taxonomy}
Existing benchmarks define their own ad hoc set of edit categories, ranging from a handful of types in MagicBrush~\cite{zhang2024magicbrush} to twenty in AnyEdit~\cite{yu2025anyedit}, making cross-comparison difficult and leaving gaps in coverage. 
We instead derive a bottom-up taxonomy from 42,400 real-world editing requests collected from the Reddit community \texttt{r/PhotoshopRequest} via the RealEdit dataset~\cite{sushko2025realedit}. Using diversity-aware sampling, hierarchical clustering, and LLM-based instruction decomposition (see Appendix~\ref{app:taxonomy-method}), we identify recurring atomic operations and group them into 20 edit action attributes. 
We distil these into 9 canonical edit types covering the operations that can be faithfully implemented in our procedural pipeline.
These edit types can be categorised into object-level edits (Removal/Addition, Movement, Rotation, Scale, Shape, Articulation, Material) and scene-level edits (Lighting, Camera Position and Zooming).

\subsection{Acquiring assets}
\label{sec:assets}

We use Infinigen~\cite{infinigen2024indoors} to procedurally generate 1{,}000
photo-realistic indoor scenes spanning bathrooms, kitchens, dining
rooms, bedrooms, and living rooms. To supplement Infinigen's
procedural materials, we curate a material dataset of 33
categories from ambientCG. 
The full material library has many near-identical entries; therefore 
we apply a greedy
furthest-point sampling algorithm over
DINOv3~\cite{simeoni2025dinov3} features to select a
visually diverse subset.

\subsection{Procedural edit generation}
\label{sec:edit_generation}

We generate image editing pairs by programmatically manipulating
Infinigen scene parameters. 
We first sample valid scene parameters and edits via rejection sampling, and then produce 2D rendered images using Blender's ray-tracing engine at 1920$\times$1080 resolution.

\mypar{View selection.}
Each scene is parsed into a structured scene graph. For each
object, we place a virtual camera on a cylinder centred at the
object's centroid with randomised distance, azimuth, and height. A
view is accepted only if the target object's visibility exceeds a
threshold $\tau_{\text{vis}} = 0.2$, estimated via face-level ray
casting. This constraint ensures that every test case clearly presents the edited object. We aim to test a VLM judge's perception, not its ability to find a needle in a haystack. Within each accepted view, we identify all visible
objects and sample edits across all enabled edit types, resetting the scene state between edits to ensure independence.

\mypar{Edit types and saved scene parameters.}
Fig.~\ref{fig:pipeline} illustrates the edit operations supported by \benchmark. Each edit is generated by intervening on a known Blender scene parameter, shown in the centre of the figure. Object-level edits modify a single target object, while scene-level edits modify global rendering or camera parameters.
\textbf{Movement} changes the object position $\mathbf{x}$ on its supporting surface, with direction quantised into eight camera-relative directions. \textbf{Rotation} changes the object orientation $\mathbf{r}$ around the vertical axis. \textbf{Scale} applies a uniform change to object scale $\mathbf{s}$ with collision checking. \textbf{Shape} modifies a continuous procedural shape parameter $\theta$, such as chair back height or table thickness. \textbf{Articulation} modifies a joint parameter $\psi$, such as opening or closing a door. \textbf{Material} changes the material assignment $m$, either for the full object or for a semantically meaningful material slot, such as an oven body or glass panel. \textbf{Removal} changes object visibility $v$ by hiding a visible object; \textbf{Addition} is obtained by swapping the before/after pair of a removal edit.
\textbf{Lighting} changes the illumination state $\ell$ by applying one of ten lighting presets. \textbf{Camera} edits modify the camera state $\mathcal{C}$, either by perturbing focal length for zoom edits or by re-sampling the camera pose on the viewing cylinder for view edits.

All edit types use rejection sampling to enforce physical plausibility. Collision checks prevent interpenetration, visibility checks ensure that the edited content remains in view, and magnitude thresholds remove negligible changes. Crucially, every edit is produced by a known procedural intervention, and each example stores the ground-truth parameter values before and after the edit. We use these parameters later to audit whether VLM judges are sensitive to the semantic correctness of an edit or merely to the visual salience induced by the underlying parameter change.

\begin{figure}[t]
  \centering
  \includegraphics[width=\textwidth]{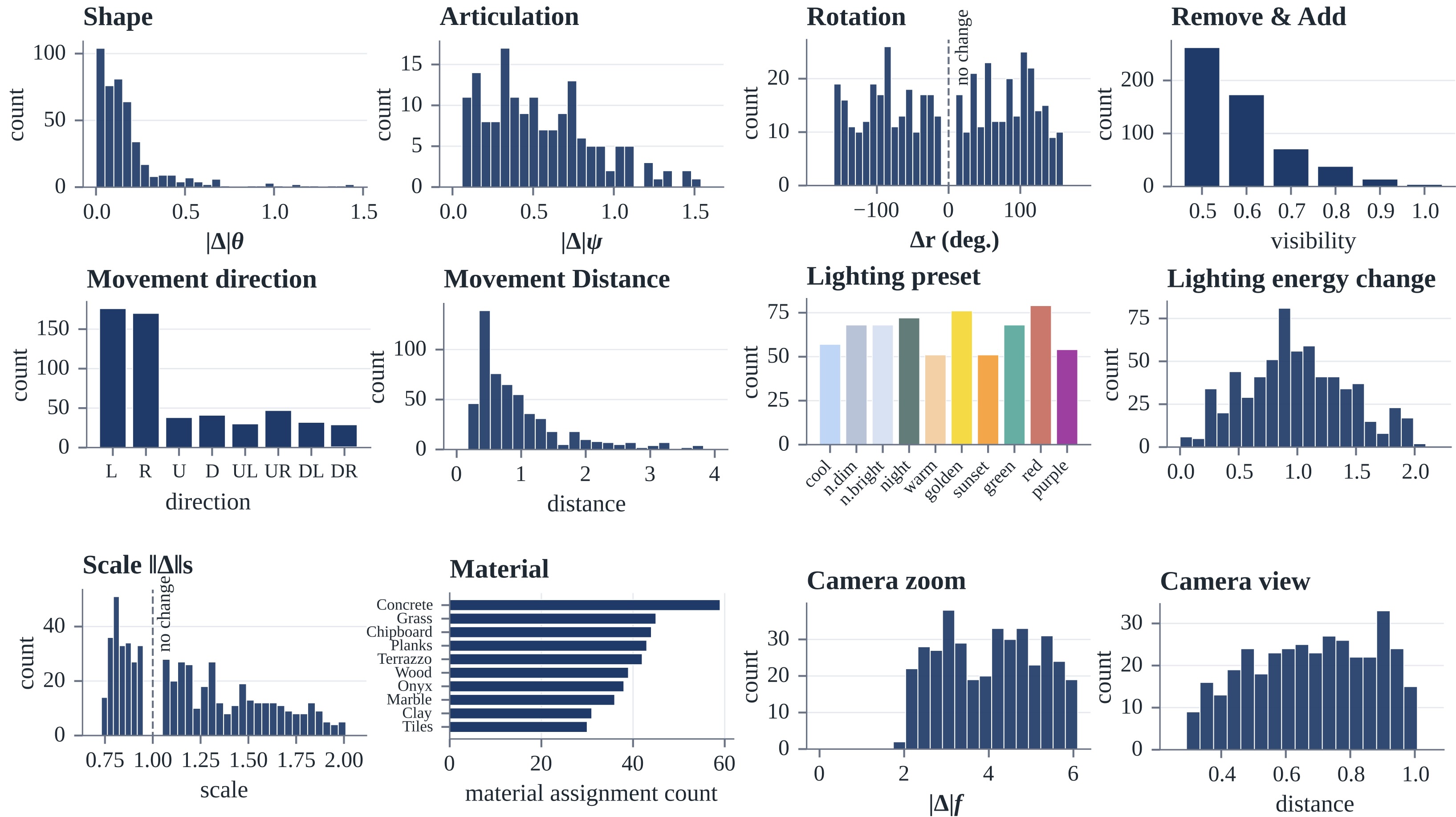}
  \caption{\textbf{Sampled edit parameter distributions.} Each plot shows the distribution of parameters sampled during edit generation. Shape, articulation, and movement distributions are clipped at the 98th percentile for readability. Dashed lines mark no-change values for rotation and scale.}
  \label{fig:edit-distributions}
\end{figure}

\mypar{Generation pool statistics.} We render 5,284 edits across 1000 3D scenes, with balanced sampling across edit types, object classes, and scenes. Fig.~\ref{fig:examples} shows edit examples, and Fig.~\ref{fig:edit-distributions} shows per-type parameter distributions.

\mypar{Counterfactual Distractors.} A judge is unreliable if it can merely detect that something changed without understanding the edit. To test this ability, we generate four types of counterfactual instructions that should be answered \emph{negatively} given the same image pair, each targeting a distinct failure mode.
\textbf{Directional:} The core edit parameter is logically inverted (e.g., left to right, larger to smaller, opened to closed), testing whether the judge perceives the direction or state of the change.
\textbf{Wrong-Object:} The edit description is retained, but the target object is replaced with a different visible object in the scene (e.g., ``Move the \emph{table} to the right'' when the chair was moved), testing object grounding.
\textbf{Cross-Edit-Type:} The instruction describes a different edit type with a visually confusable effect (e.g., camera zoom paired with object scaling), testing whether the judge identifies what kind of change occurred.
\textbf{No edit:} The instruction specifies that nothing should be changed, testing open-ended visual differencing.

\mypar{\benchmark{}.}
For evaluation (Sec.~\ref{sec:experiments}), we subsample from the larger pool of 5,284 rendered edits to obtain balanced coverage across edit types, object classes, and scenes, and to keep evaluation cost across 25 judge configurations manageable. We obtain a balanced core benchmark of 1,500 source/edited image pairs, corresponding to 3,000 rendered images. Each image pair includes one factual instruction and a set of counterfactual distractors, producing 7,318 verification triplets in total: 1,500 factual and 5,818 counterfactual. 

\mypar{Manual curation.}
Procedural rendering occasionally produces pairs that cannot be verified reliably from the images alone. We therefore manually reviewed all 1,500 pairs and removed those where (i)~the instruction text did not precisely match the rendered edit, or (ii)~the change was too subtle to be reliably perceived. This removes 118 pairs (7.9\%), yielding a curated subset of 1,382 pairs, on which we report all main results.

\section{Experimental results}
\label{sec:experiments}

Using \benchmark, we run controlled experiments with recent VLM judges to isolate individual factors affecting judge accuracy. We establish the evaluation task, validate data quality, and define the prominent VLM paradigms evaluated to establish baseline performance on our benchmark.

\mypar{Evaluation protocol.}
We frame edit verification as a binary classification of an edit triplet: the source image, the edited image, and the text instruction. Positive triplets combine a source image, an edited image, and the corresponding factual instruction. Counterfactual triplets use the same source--edited image pair, but replace the factual instruction with an incorrect distractor described in Sec.~\ref{sec:edit_generation}. We report AUROC over the binary labels and predicted continuous scores, measuring whether judges assign higher scores to factual triplets than to distractor triplets.

As a stricter complementary diagnostic, we additionally report MC-style forced-choice accuracy, which counts an example as correct when the judge assigns the unique highest score to the factual instruction among its matched counterfactual distractors. If the factual instruction ties for first among $k$ candidates, it receives $1/k$ credit.

\subsection{VLM-as-a-judge methods}
\label{sec:baselines}

To evaluate VLM-as-a-judge methods systematically, we group existing approaches into three scoring paradigms and evaluate representative models from each.

\textit{Generative scoring} instructs a base VLM to generate a numerical score conditioned on the edit triplet, as in VIE-score. Since our benchmark measures only edit correctness, we use only the correctness-related prompt of each protocol (e.g., semantic consistency in VIE-score, instruction following in KRIS) and omit separate quality prompts. Prior work often couples such templates to proprietary models like GPT-4o; to separate template effects from base-model effects, we evaluate the same templates across multiple base models.

\textit{Probabilistic scoring} uses the model's next-token distribution rather than a sampled numerical response. We include a simple binary log-probability baseline, where the model is prompted to classify each triplet as \texttt{True} or \texttt{False}. We use the next-token probability assigned to \texttt{True} as a continuous correctness score. This pointwise scoring formulation is closely related to the inference mechanism of Qwen3-VL-Reranker~\cite{li2026qwen3}, which we also include for comparison.

\textit{Supervised scoring} uses models trained in prior work~\cite{ye2025imgedit, wu2025editreward,luo2025editscore} using supervised fine-tuning or reinforcement learning to align edit-quality scores with human preferences.

\subsection{Main results}

Table~\ref{tab:evaluation_results} reports AUROC across nine edit categories for 25 judge configurations grouped by scoring paradigm.
First, we observe that the widely adopted VIEScore template is unreliable for the model that popularised it. VIEScore was introduced with GPT-4o, where it reaches only 69.0\% AUROC on \benchmark. Holding the template fixed, performance varies widely across base models (51.5 with Qwen3-VL 4B to 88.5 with GPT-5.6-sol), so a popular template does not by itself imply reliable verification. Only the latest frontier models (GPT-5.6-sol, 88.5; Opus 5, 83.5) are consistently strong judges. However benchmarks rely on older model versions that prove to be unreliable.
Second, open-source models with simple probabilistic scoring match or exceed earlier-generation proprietary models, though frontier models remain ahead. Holding the base model fixed, the scoring formulation matters substantially. Gemma-4 reaches 74.3 with VIEScore, 74.6 with ImgEdit prompting, and 81.0 with binary log-prob scoring, which is the best open-source result, above GPT-5.4 (77.9), GPT-4o and Gemini-2.5-flash. Probabilistic scoring is the strongest zero-shot formulation for every open-source backbone where we evaluate it. This indicates that mainstream zero-shot templates underutilise the underlying VLMs.
Third, supervised scoring judges trained on human preferences do not consistently transfer to factual verification. ImgEdit-Judge reaches 51.7\% (i.e., near chance performance), EditScore 63.4\%, and EditReward 66.3\%. None outperforms simple log-prob scoring on its own base model (65.4 for Qwen3-VL 4B; 70.1 for Qwen2.5-VL 7B). Preference tuning evidently captures aggregate quality cues that do not generalise to whether the requested edit was executed.
Fourth, the performance is uneven across edit categories. Removal (98.8\% best) and material (90.1\%, reached by open-source Gemma-4 with log-prob scoring) are detected reliably, while geometric and spatial edits remain weak: even the best judge reaches only 83.0 on shape and 80.1 on rotation, and no open-source configuration exceeds 62.5 on shape or 63.8 on rotation.

To validate that our synthetic edits are perceptually unambiguous, we construct a multiple-choice human study (more details in Appendix~\ref{app:human}). Five participants replied to 100 multiple choice (MC) questions with counterfactual distractors and achieved 95.2\% accuracy on average.

Under the analogous MC-style forced-choice diagnostic, however, the best VLM judge reaches only 71.5\% (Tab.~\ref{tab:evaluation_results}). This gap suggests that model failures are unlikely to stem from ambiguous synthetic examples, but rather from limitations in the judges' ability to discriminate between factual and counterfactual edit triplets.

\begin{table}[t]
\centering
\caption{\textbf{AUROC ($\uparrow$) of VLM-as-a-judge methods on the curated \benchmark{}}
\textbf{Bold}: best, \underline{underline}: second best among open models; proprietary models are shown for reference.
MC Acc.: multiple-choice accuracy of ranking the correct instruction above all wrong ones for the same edit, with ties split evenly (chance $\approx$ 21\%).}
\label{tab:evaluation_results}
\footnotesize
\setlength{\tabcolsep}{3.5pt}
\setlength{\aboverulesep}{0pt}
\setlength{\belowrulesep}{0pt}
\begin{adjustbox}{max width=\linewidth}
\NiceTab{l@{\hspace{6pt}}l ccccccccc c c}
\toprule
\textbf{Method} & \textbf{Base Model} & \textbf{Rem} & \textbf{Mat} & \textbf{Lgt} & \textbf{Cam} & \textbf{Mov} & \textbf{Scl} & \textbf{Shp} & \textbf{Art} & \textbf{Rot} & \textbf{Overall} & \textbf{MC Acc.} \\
\midrule
\tbparadigmC{genGray}{Generative scoring} \\
\rGen \hspace{0.1em} VIE-score~\cite{ku2024viescore} & \llava\tbgap LLaVA-OV-7B & 58.8 & 57.1 & 55.0 & 48.1 & 48.6 & 49.1 & 39.4 & 53.2 & 57.8 & 51.9 & 21.9 \\
\tbdots{genGray}
\rGen & \qwen\tbgap Qwen3-VL-4B & 76.8 & 61.2 & 63.1 & 43.8 & 47.8 & 45.2 & 36.3 & 54.1 & 34.7 & 51.5 & 13.8 \\
\tbdots{genGray}
\rGen & \qwen\tbgap Qwen3-VL-8B & 75.9 & 68.0 & 66.1 & 47.5 & 56.3 & 54.0 & 51.0 & 57.0 & 33.4 & 56.6 & 18.3 \\
\tbdots{genGray}
\rGen & \qwen\tbgap Qwen3-VL-8B + Think & 85.0 & 76.6 & 73.1 & 49.7 & 63.7 & 49.9 & 56.2 & 76.4 & 46.6 & 64.1 & 30.8 \\
\tbdots{genGray}
\rGen & \qwen\tbgap Qwen3-VL-32B & 85.5 & 80.3 & 74.9 & 60.4 & 69.9 & 55.9 & 48.2 & 71.7 & 40.6 & 65.3 & 32.3 \\
\tbdots{genGray}
\rGen & \qwen\tbgap Qwen3-VL-32B + Think & 91.4 & 84.1 & 82.4 & 69.3 & 69.2 & 57.5 & 56.4 & \textbf{79.1} & 47.3 & 70.7 & 41.1 \\
\tbdots{genGray}
\rGen & \gemma\tbgap Gemma-4-31B & 89.9 & 82.2 & 83.0 & 72.4 & 83.6 & 66.5 & 59.1 & 70.5 & \underline{61.5} & 74.3 & 45.0 \\
\tbsepc{genGray}
\rGen \hspace{0.1em} Kris~\cite{wu2025krisbench} & \qwen\tbgap Qwen3-VL-8B & 84.7 & 78.7 & 79.4 & 57.5 & 69.5 & 61.2 & 52.3 & 70.0 & 37.8 & 65.7 & 34.8 \\
\tbdots{genGray}
\rGen & \gemma\tbgap Gemma-4-31B & 92.8 & 81.1 & 79.1 & 73.4 & 83.1 & \underline{67.2} & 59.8 & 65.2 & 54.6 & 72.9 & 46.4 \\
\tbsepc{genGray}
\rGen \hspace{0.1em} ImgEdit~\cite{ye2025imgedit} & \qwen\tbgap Qwen3-VL-8B & 80.4 & 73.2 & 74.1 & 62.0 & 64.7 & 60.4 & 49.0 & 64.7 & 37.9 & 62.9 & 29.8 \\
\tbdots{genGray}
\rGen & \gemma\tbgap Gemma-4-31B & 90.7 & 82.3 & \underline{86.2} & \underline{79.4} & \underline{83.8} & 64.5 & \textbf{62.5} & 66.7 & 55.5 & \underline{74.6} & 46.8 \\
\tbparadigmC{propGray}{\color{black!70}Generative scoring (proprietary models)}  \\
\rProp \hspace{0.1em} VIE-score & \faded{\gemini}\tbgap Gemini-2.5-flash & 94.6 & 86.2 & 83.8 & 71.8 & 76.5 & 57.1 & 58.2 & 72.8 & 54.0 & 72.8 & 44.6 \\
\tbdots{propGray}
\rProp & \faded{\openai}\tbgap GPT-4.1-mini & 80.1 & 81.1 & 79.8 & 59.2 & 71.3 & 57.8 & 54.9 & 67.2 & 55.0 & 67.4 & 32.7 \\
\tbdots{propGray}
\rProp & \faded{\openai}\tbgap GPT-4o & 89.2 & 84.6 & 77.6 & 60.0 & 74.5 & 56.2 & 54.5 & 70.9 & 53.7 & 69.0 & 35.8 \\
\tbdots{propGray}
\rProp & \faded{\openai}\tbgap GPT-5.4 & 95.6 & 83.6 & 88.2 & 71.2 & 85.1 & 73.2 & 66.0 & 74.1 & 64.4 & 77.9 & 49.5 \\
\tbdots{propGray}
\rProp & \faded{\openai}\tbgap GPT-5.6-sol & 97.8 & 88.2 & 91.4 & 95.2 & 95.0 & 84.4 & 83.0 & 81.2 & 80.1 & 88.5 & 71.5 \\
\tbdots{propGray}
\rProp & \faded{\claude}\tbgap Opus 5 & 98.8 & 88.3 & 86.0 & 92.7 & 91.8 & 76.8 & 73.1 & 72.9 & 70.8 & 83.5 & 61.9 \\
\tbparadigmC{probGray}{Probabilistic scoring} \\

\rProb \hspace{0.1em} Qwen-Reranker & \qwen\tbgap Qwen3-VL-Reranker-8B & \underline{96.2} & 77.8 & 81.8 & 60.4 & 64.3 & 50.6 & 46.9 & 75.2 & 51.1 & 67.1 & 43.5 \\
\tbsepc{probGray}
\rProb \hspace{0.1em} Log-prob & \qwen\tbgap Qwen3-VL-4B & 92.3 & 82.9 & 78.3 & 64.9 & 64.2 & 47.7 & 42.4 & \underline{76.8} & 39.4 & 65.4 & 41.9 \\
\tbdots{probGray}
\rProb & \qwen\tbgap Qwen3-VL-8B & \underline{96.2} & \underline{84.2} & 82.7 & 78.4 & 80.5 & 56.5 & 42.6 & 73.6 & 40.6 & 70.6 & \underline{48.3} \\
\tbdots{probGray}
\rProb & \qwen\tbgap Qwen2.5-VL-7B & 93.9 & 82.0 & 84.3 & 73.9 & 73.9 & 60.1 & 40.6 & 76.1 & 46.1 & 70.1 & 45.4 \\
\tbdots{probGray}
\rProb & \gemma\tbgap Gemma-4-31B & \textbf{96.3} & \textbf{90.1} & \textbf{89.3} & \textbf{88.7} & \textbf{91.6} & \textbf{73.0} & \underline{62.2} & 74.1 & \textbf{63.8} & \textbf{81.0} & \textbf{60.8} \\

\tbparadigmC{supGray}{Supervised scoring} \\
\rSup \hspace{0.1em} EditScore~\cite{luo2025editscore} & \qwen\tbgap Qwen3-VL-4B & 88.3 & 81.0 & 74.1 & 47.6 & 65.2 & 56.0 & 50.9 & 59.5 & 48.0 & 63.4 & 37.2 \\
\tbsepc{supGray}
\rSup \hspace{0.1em} ImgEdit-Judge~\cite{ye2025imgedit} & \qwen\tbgap Qwen2.5-VL-7B & 57.4 & 59.5 & 45.1 & 50.7 & 58.6 & 42.9 & 52.6 & 60.2 & 38.1 & 51.7 & 19.5 \\
\tbsepc{supGray}
\rSup \hspace{0.1em} EditReward~\cite{wu2025editreward} & \qwen\tbgap Qwen2.5-VL-7B & 91.0 & 74.2 & 79.1 & 66.2 & 70.2 & 57.1 & 45.6 & 62.4 & 51.1 & 66.3 & 40.0 \\

\bottomrule
\end{NiceTabular}
\end{adjustbox}
\vspace{0pt}

\raggedright\scriptsize
Rem = removal, Mat = material, Lgt = light, Cam = camera, Mov = movement, Scl = scale, Shp = shape, Art = articulation, Rot = rotation.
\vspace{-12pt}
\end{table}

\subsection{Correlation with human-preference benchmarks}
\label{sec:exp_correlation_human}

Here, we ask whether performance on \benchmark{} agrees with existing human-preference edit benchmarks. For GenAI-Bench~\cite{jiang2024genai} and EditRewardBench~\cite{wu2025editreward}, we measure whether each judge assigns a higher score to the human-preferred candidate, and compare this preference accuracy with the judge's macro edit-type AUROC on \benchmark. Across 11 judge configurations, macro AUROC on \benchmark{} is strongly Spearman rank-correlated with GenAI-Bench ($\rho=0.95$) and EditRewardBench ($\rho=0.71$). As a reference point, the two human-preference benchmarks correlate with each other at $\rho=0.83$. This suggests that \benchmark{} captures judge-level differences relevant to real-image preference evaluation, while offering controlled ground truth and edit-type-level diagnostics unavailable in preference-only datasets (per-judge results in Appendix~\ref{app:preference-alignment}).

\subsection{Analysis}
\label{sec:analysis}

\mypar{Image placement.}
Prior VIE-score implementations use different multimodal prompt orderings across model providers. Images are placed before the text instructions for Gemini, but after the text for GPT models. We ablate this choice by evaluating both orderings for Gemini-2.5-flash and GPT-4.1-mini. As shown in Appendix~\ref{app:image-placement}, placing images after the instruction improves AUROC by up to 14.1 points (13.7 points for Gemini-2.5-flash and 14.1 points for GPT-4.1-mini). This shows that prompt formatting choices, often treated as implementation details, can materially change judge rankings and motivate fixing the prompt format across baselines.

\begin{figure}[t]
  \centering
  \includegraphics[width=\textwidth]{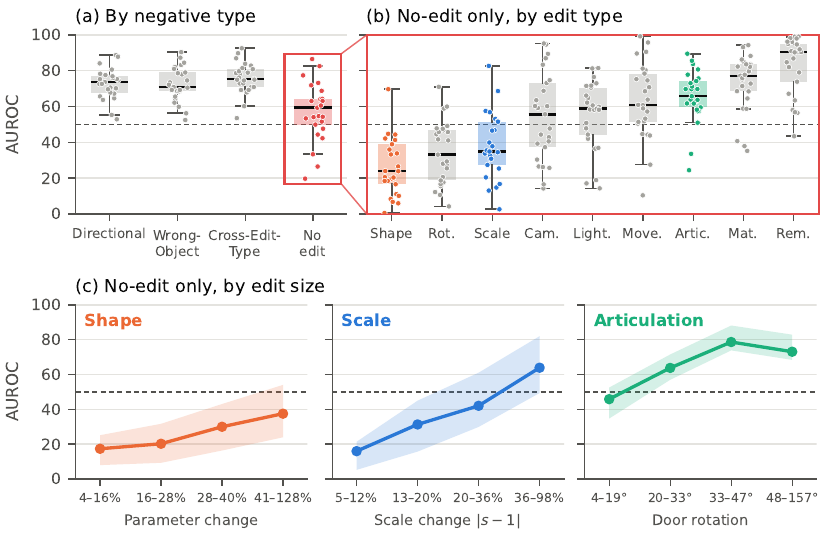}
  \caption{\textbf{No-edit instructions are the main failure mode, driven by small edits.}
  (a) AUROC of factual triplets against each counterfactual type; one dot per judge (25 configurations).
  (b) No-edit counterfactuals only, per edit type.
  (c) No-edit AUROC per quartile of ground-truth edit magnitude for shape, scale, and articulation;
  lines show the mean across judges, bands the interquartile range. Dashed line: chance.}
  \label{fig:no_edit}
\end{figure}

\mypar{No-edit counterfactuals are the main failure mode.}
We first break down performance by counterfactual type (Fig.~\ref{fig:no_edit}a). Directional, wrong-object, and cross-type counterfactuals are all rejected above chance, but far from perfectly, with a median AUROC in the low-to-mid 70s. No-edit counterfactuals are substantially harder: the median AUROC is markedly lower and varies widely across judges, making ``nothing changed'' the most difficult counterfactual to reject. Splitting no-edit counterfactuals by edit type (Fig.~\ref{fig:no_edit}b) shows that this aggregate hides two distinct regimes. For removal and material edits, judges usually identify the change. For shape, rotation, and scale edits, the median AUROC falls well below chance, meaning judges systematically rate the no-edit instruction as \emph{more} consistent with the image pair than the true edit. Camera, move, lighting, and articulation edits fall in between.

To test whether this blind spot depends on edit magnitude, we stratify factual/no-edit comparisons by ground-truth Blender parameter changes (Fig.~\ref{fig:no_edit}c). We observe a consistent salience effect where AUROC rises with edit magnitude across shape, scale, and articulation. Scale moves from well below chance to above chance as changes grow, while articulation crosses chance above roughly $20^\circ$ of door rotation. Shape edits remain below chance even in the top quartile, suggesting geometric changes are difficult to perceive at any tested magnitude. Overall, the no-edit blind spot reflects a limitation of visual differencing under low salience rather than a fixed failure of the prompt.

\paragraph{Recommendations for judge validation.}
Aligning with human preferences is a necessary but insufficient test for an edit judge: models that match GenAI-Bench rankings still overlook spatial errors and tolerate unrequested changes. We therefore recommend evaluating \emph{correctness} alongside preference using controlled benchmarks like \benchmark. Specifically, we advise practitioners to: 
(i)~report per-edit-type reliability, as aggregate scores mask large performance gaps (Tab.~\ref{tab:evaluation_results}); 
(ii)~test sensitivity to unrequested changes (Fig.~\ref{fig:no_edit}); 
(iii)~standardize and report the prompt format, which alone shifts AUROC by up to 14.1 points; and 
(iv)~include a binary log-prob baseline, the strongest zero-shot formulation on every open-source backbone where we tested it.

\section{Conclusions}
\label{sec:conclusions}

We introduced \benchmark, a diagnostic benchmark of controlled image editing pairs to audit VLM-based judges against known ground truth. Auditing 25 judge configurations across zero-shot prompting, probabilistic scoring, and fine-tuned reward models reveals that current protocols are far less reliable than their adoption suggests.
\benchmark also exposes failure modes, like weak open-ended visual differencing, that aggregate human-preference correlations miss. Despite using synthetic data, our judge rankings correlate strongly with real-image human preferences ($\rho$=0.95 with GenAI-Bench), showing that controlled audits capture real-world failures. Ultimately, we hope \benchmark encourages the field to report per-edit-type reliability, validate judges using no-edit and counterfactual pairs, and standardise multimodal prompt formatting.

\mypar{Limitations.} \benchmark renders indoor scenes only, edits are produced by deterministic 3D interventions rather than by editing models, and binary verification is a targeted probe rather than a substitute for holistic edit-quality judgment. We elaborate more in Appendix~\ref{app:limitations}.

\mypar{Acknowledgments.} D. Papadopoulos and M. Schouten were supported by
the DFF Sapere Aude Starting Grant “ACHILLES”. J. Parslov was supported by Danish Data Science Academy, which is funded by the Novo Nordisk Foundation (NNF21SA0069429) and VILLUM FONDEN (40516)

\bibliographystyle{splncs04}
\bibliography{main}

\begin{thebibliography}{10}
\providecommand{\url}[1]{\texttt{#1}}
\providecommand{\urlprefix}{URL }
\providecommand{\doi}[1]{https://doi.org/#1}

\bibitem{Blender2026}
{Blender Online Community}: Blender - A 3D Modelling and Rendering Package.
  Blender Foundation, Blender Institute, Amsterdam (2026)

\bibitem{brack2024ledits}
Brack, M., Friedrich, F., Kornmeier, K., Tsaban, L., Schramowski, P., Kersting,
  K., Passos, A.: Ledits++: Limitless image editing using text-to-image models.
  In: Proceedings of the IEEE/CVF Conference on Computer Vision and Pattern
  Recognition (CVPR) (2024)

\bibitem{brooks2023instructpix2pix}
Brooks, T., Holynski, A., Efros, A.A.: Instructpix2pix: Learning to follow
  image editing instructions. In: Proceedings of the IEEE/CVF Conference on
  Computer Vision and Pattern Recognition (CVPR) (2023)

\bibitem{butler2012naturalistic}
Butler, D.J., Wulff, J., Stanley, G.B., Black, M.J.: A naturalistic open source
  movie for optical flow evaluation. In: European conference on computer
  vision. pp. 611--625. Springer (2012)

\bibitem{pmlr-v235-chen24h}
Chen, D., Chen, R., Zhang, S., Wang, Y., Liu, Y., Zhou, H., Zhang, Q., Wan, Y.,
  Zhou, P., Sun, L.: {MLLM}-as-a-judge: Assessing multimodal {LLM}-as-a-judge
  with vision-language benchmark. In: Salakhutdinov, R., Kolter, Z., Heller,
  K., Weller, A., Oliver, N., Scarlett, J., Berkenkamp, F. (eds.) Proceedings
  of the 41st International Conference on Machine Learning. Proceedings of
  Machine Learning Research, vol.~235, pp. 6562--6595. PMLR (21--27 Jul 2024),
  \url{https://proceedings.mlr.press/v235/chen24h.html}

\bibitem{chen2023chatgpt}
Chen, L., Zaharia, M., Zou, J.: How is chatgpt's behavior changing over time?
  arXiv preprint arXiv:2307.09009  (2023)

\bibitem{chen2025mj}
Chen, Z., Wen, Z., Du, Y., Zhou, Y., Cui, C., Han, S., Weng, J., Wang, C.,
  Tong, Z., HUANG, L., et~al.: Mj-bench: Is your multimodal reward model really
  a good judge for text-to-image generation? Advances in Neural Information
  Processing Systems  \textbf{38} (2025)

\bibitem{dubois2023alpacafarm}
Dubois, Y., Li, C.X., Taori, R., Zhang, T., Gulrajani, I., Ba, J., Guestrin,
  C., Liang, P.S., Hashimoto, T.B.: Alpacafarm: A simulation framework for
  methods that learn from human feedback. Advances in Neural Information
  Processing Systems  \textbf{36},  30039--30069 (2023)

\bibitem{fu2024blink}
Fu, X., Hu, Y., Li, B., Feng, Y., Wang, H., Lin, X., Roth, D., Smith, N.A., Ma,
  W.C., Krishna, R.: Blink: Multimodal large language models can see but not
  perceive. In: European Conference on Computer Vision. pp. 148--166. Springer
  (2024)

\bibitem{guan2024hallusionbench}
Guan, T., Liu, F., Wu, X., Xian, R., Li, Z., Liu, X., Wang, X., Chen, L.,
  Huang, F., Yacoob, Y., et~al.: Hallusionbench: an advanced diagnostic suite
  for entangled language hallucination and visual illusion in large
  vision-language models. In: Proceedings of the IEEE/CVF conference on
  computer vision and pattern recognition. pp. 14375--14385 (2024)

\bibitem{isola2017image}
Isola, P., Zhu, J.Y., Zhou, T., Efros, A.A.: Image-to-image translation with
  conditional adversarial networks. In: Proceedings of the IEEE conference on
  computer vision and pattern recognition. pp. 1125--1134 (2017)

\bibitem{jiang2024genai}
Jiang, D., Ku, M., Li, T., Ni, Y., Sun, S., Fan, R., Chen, W.: Genai arena: An
  open evaluation platform for generative models. Advances in Neural
  Information Processing Systems  \textbf{37},  79889--79908 (2024)

\bibitem{jiang2026geditbench}
Jiang, Z., Sun, Z., Zeng, X., Yang, Y., Zhang, X., Wu, Y., Cheng, W., Yu, G.,
  Yang, X., Wen, B.: Geditbench v2: A human-aligned benchmark for general image
  editing. arXiv preprint arXiv:2603.28547  (2026)

\bibitem{johnson2017clevr}
Johnson, J., Hariharan, B., Van Der~Maaten, L., Fei-Fei, L., Lawrence~Zitnick,
  C., Girshick, R.: Clevr: A diagnostic dataset for compositional language and
  elementary visual reasoning. In: Proceedings of the IEEE conference on
  computer vision and pattern recognition. pp. 2901--2910 (2017)

\bibitem{kawar2023imagic}
Kawar, B., Zada, S., Lang, O., Tov, O., Chang, H., Dekel, T., Mosseri, I.,
  Irani, M.: Imagic: Text-based real image editing with diffusion models. In:
  Proceedings of the IEEE/CVF Conference on Computer Vision and Pattern
  Recognition (CVPR) (2023)

\bibitem{kim2026vlmsubtlebench}
Kim, M., Lee, S., Park, D.: {VLM}-subtlebench: How far are {VLM}s from
  human-level subtle comparative reasoning? In: The Fourteenth International
  Conference on Learning Representations (2026)

\bibitem{koishigarina2025clip}
Koishigarina, D., Uselis, A., Oh, S.J.: Clip behaves like a bag-of-words model
  cross-modally but not uni-modally. In: International Conference on Learning
  Representations. vol.~2026, pp. 96106--96127 (2026)

\bibitem{ku2024viescore}
Ku, M., Jiang, D., Wei, C., Yue, X., Chen, W.: Viescore: Towards explainable
  metrics for conditional image synthesis evaluation. In: Proceedings of the
  62nd Annual Meeting of the Association for Computational Linguistics (Volume
  1: Long Papers). pp. 12268--12290 (2024)

\bibitem{gu2024survey}
Li, D., Jiang, B., Huang, L., Beigi, A., Zhao, C., Tan, Z., Bhattacharjee, A.,
  Jiang, Y., Chen, C., Wu, T., et~al.: From generation to judgment:
  Opportunities and challenges of llm-as-a-judge. In: Proceedings of the 2025
  Conference on Empirical Methods in Natural Language Processing. pp.
  2757--2791 (2025)

\bibitem{li2026qwen3}
Li, M., Zhang, Y., Long, D., Chen, K., Song, S., Bai, S., Yang, Z., Xie, P.,
  Yang, A., Liu, D., et~al.: Qwen3-vl-embedding and qwen3-vl-reranker: A
  unified framework for state-of-the-art multimodal retrieval and ranking.
  arXiv preprint arXiv:2601.04720  (2026)

\bibitem{li2026genarena}
Li, R., Qu, L., Zhang, J., Gui, D., Xu, M., Zhang, X., Hu, H., Wang, W., Wang,
  J.: Genarena: How can we achieve human-aligned evaluation for visual
  generation tasks? arXiv preprint arXiv:2602.06013  (2026)

\bibitem{li2023evaluating}
Li, Y., Du, Y., Zhou, K., Wang, J., Zhao, W.X., Wen, J.R.: Evaluating object
  hallucination in large vision-language models. In: Proceedings of the 2023
  conference on empirical methods in natural language processing. pp. 292--305
  (2023)

\bibitem{li2021openrooms}
Li, Z., Yu, T.W., Sang, S., Wang, S., Song, M., Liu, Y., Yeh, Y.Y., Zhu, R.,
  Gundavarapu, N., Shi, J., et~al.: Openrooms: An open framework for
  photorealistic indoor scene datasets. In: Proceedings of the IEEE/CVF
  conference on computer vision and pattern recognition. pp. 7190--7199 (2021)

\bibitem{liu2025step1xedit}
Liu, S., et~al.: Step1x-edit: A practical framework for general image editing
  (gedit-bench). arXiv preprint arXiv:2504.17761  (2025)

\bibitem{liu2025omnidiff}
Liu, Y., Hou, S., Hou, S., Du, J., Meng, S., Huang, Y.: Omnidiff: A
  comprehensive benchmark for fine-grained image difference captioning. In:
  Proceedings of the IEEE/CVF International Conference on Computer Vision. pp.
  21440--21449 (2025)

\bibitem{luo2025editscore}
Luo, X., Wang, J., Wu, C., Xiao, S., Jiang, X., Lian, D., Zhang, J., Liu, D.,
  Liu, Z.: Editscore: Unlocking online rl for image editing via high-fidelity
  reward modeling. In: International Conference on Learning Representations.
  vol.~2026, pp. 33027--33056 (2026)

\bibitem{huang2024i2ebench}
Ma, Y., Ji, J., Ye, K., Lin, W., Wang, Z., Zheng, Y., Zhou, Q., Sun, X., Ji,
  R.: I2ebench: A comprehensive benchmark for instruction-based image editing.
  Advances in Neural Information Processing Systems  \textbf{37},  41494--41516
  (2024)

\bibitem{mayer2016large}
Mayer, N., Ilg, E., Hausser, P., Fischer, P., Cremers, D., Dosovitskiy, A.,
  Brox, T.: A large dataset to train convolutional networks for disparity,
  optical flow, and scene flow estimation. In: Proceedings of the IEEE
  conference on computer vision and pattern recognition. pp. 4040--4048 (2016)

\bibitem{michel2023object}
Michel, O., Bhattad, A., VanderBilt, E., Krishna, R., Kembhavi, A., Gupta, T.:
  Object 3dit: Language-guided 3d-aware image editing. Advances in Neural
  Information Processing Systems  \textbf{36},  3497--3516 (2023)

\bibitem{park2019robust}
Park, D.H., Darrell, T., Rohrbach, A.: Robust change captioning. In:
  Proceedings of the IEEE/CVF International Conference on Computer Vision. pp.
  4624--4633 (2019)

\bibitem{scikit-learn}
Pedregosa, F., Varoquaux, G., Gramfort, A., Michel, V., Thirion, B., Grisel,
  O., Blondel, M., Prettenhofer, P., Weiss, R., Dubourg, V., Vanderplas, J.,
  Passos, A., Cournapeau, D., Brucher, M., Perrot, M., Duchesnay, E.:
  Scikit-learn: Machine learning in {P}ython. Journal of Machine Learning
  Research  \textbf{12},  2825--2830 (2011)

\bibitem{pozzobon2023challenges}
Pozzobon, L., Ermis, B., Lewis, P., Hooker, S.: On the challenges of using
  black-box apis for toxicity evaluation in research. In: Proceedings of the
  2023 conference on empirical methods in natural language processing. pp.
  7595--7609 (2023)

\bibitem{qian2025giebench}
Qian, Y., Lu, J., Fu, T.J., Wang, X., Chen, C., Yang, Y., Hu, W., Gan, Z.:
  Gie-bench: Towards grounded evaluation for text-guided image editing. arXiv
  preprint arXiv:2505.11493  (2025)

\bibitem{qu2025ie}
Qu, B., Sun, S., Liang, X., Gao, W.: Ie-critic-r1: Advancing the explanatory
  measurement of text-driven image editing for human perception alignment.
  arXiv preprint arXiv:2511.18055  (2025)

\bibitem{radford2021learning}
Radford, A., Kim, J.W., Hallacy, C., Ramesh, A., Goh, G., Agarwal, S., Sastry,
  G., Askell, A., Mishkin, P., Clark, J., et~al.: Learning transferable visual
  models from natural language supervision. In: International conference on
  machine learning. pp. 8748--8763. PmLR (2021)

\bibitem{rahmanzadehgervi2024vision}
Rahmanzadehgervi, P., Bolton, L., Taesiri, M.R., Nguyen, A.T.: Vision language
  models are blind. In: Proceedings of the Asian Conference on Computer Vision.
  pp. 18--34 (2024)

\bibitem{infinigen2024indoors}
Raistrick, A., Mei, L., Kayan, K., Yan, D., Zuo, Y., Han, B., Wen, H., Parakh,
  M., Alexandropoulos, S., Lipson, L., Ma, Z., Deng, J.: Infinigen indoors:
  Photorealistic indoor scenes using procedural generation. In: Proceedings of
  the IEEE/CVF Conference on Computer Vision and Pattern Recognition (CVPR).
  pp. 21783--21794 (June 2024)

\bibitem{sheynin2024emuedit}
Sheynin, S., Polyak, A., Singer, U., Kirstain, Y., Zohar, A., Ashual, O.,
  Parikh, D., Taigman, Y.: Emu edit: Precise image editing via recognition and
  generation tasks. In: Proceedings of the IEEE/CVF Conference on Computer
  Vision and Pattern Recognition (CVPR) (2024)

\bibitem{simeoni2025dinov3}
Sim{\'e}oni, O., Vo, H.V., Seitzer, M., Baldassarre, F., Oquab, M., Jose, C.,
  Khalidov, V., Szafraniec, M., Yi, S., Ramamonjisoa, M., et~al.: Dinov3. arXiv
  preprint arXiv:2508.10104  (2025)

\bibitem{sun2025iebench}
Sun, S., Qu, B., Liang, X., Fan, S., Gao, W.: Ie-bench: Advancing the
  measurement of text-driven image editing for human perception alignment.
  arXiv preprint arXiv:2501.09927  (2025)

\bibitem{sushko2025realedit}
Sushko, P., Bharadwaj, A., Lim, Z.Y., Ilin, V., Caffee, B., Chen, D., Salehi,
  M., Hsieh, C.Y., Krishna, R.: Realedit: Reddit edits as a large-scale
  empirical dataset for image transformations. In: Proceedings of the Computer
  Vision and Pattern Recognition Conference. pp. 13403--13413 (2025)

\bibitem{tan2024judgebench}
Tan, S., Zhuang, S., Montgomery, K., Tang, W., Cuadron, A., Wang, C., Popa, R.,
  Stoica, I.: Judgebench: A benchmark for evaluating llm-based judges. In:
  International Conference on Learning Representations. vol.~2025, pp.
  63277--63303 (2025)

\bibitem{tong2024eyes}
Tong, S., Liu, Z., Zhai, Y., Ma, Y., LeCun, Y., Xie, S.: Eyes wide shut?
  exploring the visual shortcomings of multimodal llms. In: 2024 IEEE/CVF
  Conference on Computer Vision and Pattern Recognition (CVPR). pp. 9568--9578.
  IEEE (2024)

\bibitem{wan2025motionedit}
Wan, Y., Ke, L., Yu, W., Chang, K.W., Yu, D.: Motionedit: Benchmarking and
  learning motion-centric image editing. arXiv preprint arXiv:2512.10284
  (2025)

\bibitem{wang2025i2i}
Wang, J., Wang, J., Duan, H., Kang, J., Zhai, G., Min, X.: I2i-bench: A
  comprehensive benchmark suite for image-to-image editing models. In:
  Proceedings of the IEEE/CVF Conference on Computer Vision and Pattern
  Recognition. pp. 15354--15364 (2026)

\bibitem{wu2025editreward}
Wu, K., Jiang, S., Ku, M., Nie, P., Liu, M., Chen, W.: Editreward: A
  human-aligned reward model for instruction-guided image editing. In: The
  Fourteenth International Conference on Learning Representations (2026)

\bibitem{wu2025krisbench}
Wu, Y., Li, Z., Hu, X., Ye, X., Zeng, X., Yu, G., Zhu, W., Schiele, B., Yang,
  M.H., Yang, X.: Kris-bench: Benchmarking next-level intelligent image editing
  models. In: Advances in Neural Information Processing Systems (NeurIPS)
  (2025)

\bibitem{yasunaga2025multimodal}
Yasunaga, M., Zettlemoyer, L., Ghazvininejad, M.: Multimodal rewardbench:
  Holistic evaluation of reward models for vision language models. arXiv
  preprint arXiv:2502.14191  (2025)

\bibitem{ye2025imgedit}
Ye, Y., He, X., Li, Z., Lin, B., Yuan, S., Yan, Z., Hou, B., Yuan, L.:
  {ImgEdit}: A unified image editing dataset and benchmark. In: Advances in
  Neural Information Processing Systems. vol.~38 (2025)

\bibitem{yosef2025editinspector}
Yosef, R., Bitton, Y., Lischinski, D., Yanuka, M.: Editinspector: A benchmark
  for evaluation of text-guided image edits. In: Proceedings of the 63rd Annual
  Meeting of the Association for Computational Linguistics (Volume 1: Long
  Papers). pp. 29503--29530 (2025)

\bibitem{yu2025anyedit}
Yu, Q., Chow, W., Yue, Z., Pan, K., Wu, Y., Wan, X., Li, J., Tang, S., Zhang,
  H., Zhuang, Y.: Anyedit: Mastering unified high-quality image editing for any
  idea. In: Proceedings of the Computer Vision and Pattern Recognition
  Conference. pp. 26125--26135 (2025)

\bibitem{yuksekgonul2022and}
Yuksekgonul, M., Bianchi, F., Kalluri, P., Jurafsky, D., Zou, J.: When and why
  vision-language models behave like bags-of-words, and what to do about it?
  In: International Conference on Learning Representations (2023)

\bibitem{zeng2023evaluating}
Zeng, Z., Yu, J., Gao, T., Meng, Y., Goyal, T., Chen, D.: Evaluating large
  language models at evaluating instruction following. In: International
  Conference on Learning Representations. vol.~2024, pp. 40193--40219 (2024)

\bibitem{zhang2024magicbrush}
Zhang, K., Mo, L., Chen, W., Sun, H., Su, Y.: Magicbrush: A manually annotated
  dataset for instruction-guided image editing. In: Oh, A., Naumann, T.,
  Globerson, A., Saenko, K., Hardt, M., Levine, S. (eds.) Advances in Neural
  Information Processing Systems. vol.~36, pp. 31428--31449. Curran Associates,
  Inc. (2023)

\bibitem{zhang2025qwen3}
Zhang, Y., Li, M., Long, D., Zhang, X., Lin, H., Yang, B., Xie, P., Yang, A.,
  Liu, D., Lin, J., et~al.: Qwen3 embedding: Advancing text embedding and
  reranking through foundation models. arXiv preprint arXiv:2506.05176  (2025)

\bibitem{zhao2024ultraedit}
Zhao, H., Ma, X.S., Chen, L., Si, S., Wu, R., An, K., Yu, P., Zhang, M., Li,
  Q., Chang, B.: Ultraedit: Instruction-based fine-grained image editing at
  scale. Advances in Neural Information Processing Systems  \textbf{37},
  3058--3093 (2024)

\bibitem{zhao2025envisioning}
Zhao, X., Zhang, P., Tang, K., Zhu, X., Li, H., Chai, W., Zhang, Z., Xia, R.,
  Zhai, G., Yan, J., et~al.: Envisioning beyond the pixels: Benchmarking
  reasoning-informed visual editing. Advances in Neural Information Processing
  Systems  \textbf{38} (2025)

\bibitem{zheng2023judging}
Zheng, L., Chiang, W.L., Sheng, Y., Zhuang, S., Wu, Z., Zhuang, Y., Lin, Z.,
  Li, Z., Li, D., Xing, E., et~al.: Judging llm-as-a-judge with mt-bench and
  chatbot arena. Advances in neural information processing systems
  \textbf{36},  46595--46623 (2023)

\bibitem{zheng2023pointodyssey}
Zheng, Y., Harley, A.W., Shen, B., Wetzstein, G., Guibas, L.J.: Pointodyssey: A
  large-scale synthetic dataset for long-term point tracking. In: Proceedings
  of the IEEE/CVF International Conference on Computer Vision. pp. 19855--19865
  (2023)

\end{thebibliography}
\newpage
\appendix
\section{Appendix - Results}
\begin{table}[t]
\centering
\caption{\textbf{AUROC ($\uparrow$) by counterfactual type for all main-table judges.}
Directional, cross-type, wrong-object, and no-edit counterfactuals are evaluated against the same positive examples; Sem.\ is the mean of Dir., Cross, and Wrong obj.
\textbf{Bold}: best, \underline{underline}: second best among open models; proprietary models are shown for reference.}
\label{tab:appendix_per_negative_auc}
\footnotesize
\setlength{\tabcolsep}{3.5pt}
\setlength{\aboverulesep}{0pt}
\setlength{\belowrulesep}{0pt}
\begin{adjustbox}{max width=\linewidth}
\renewcommand{\tbdots}[1]{\tbdotcline{2-7}}
\NiceTab{l@{\hspace{6pt}}l ccccc}
\toprule
\textbf{Method} & \textbf{Base Model} & \textbf{Dir.} & \textbf{Cross} & \textbf{Wrong obj.} & \textbf{No edit} & \textbf{Sem.} \\
\midrule
\tbparadigmType{genGray}{Generative scoring} \\
\rGen \hspace{0.1em} VIE-score~\cite{ku2024viescore} & \llava\tbgap LLaVA-OV-7B & 52.9 & 53.5 & 52.5 & 51.6 & 53.0 \\
\tbdots{genGray}
\rGen & \qwen\tbgap Qwen3-VL-4B & 65.7 & 67.9 & 56.2 & 19.7 & 63.3 \\
\tbdots{genGray}
\rGen & \qwen\tbgap Qwen3-VL-8B & 67.2 & 69.2 & 62.2 & 26.5 & 66.2 \\
\tbdots{genGray}
\rGen & \qwen\tbgap Qwen3-VL-8B + Think & 63.7 & 69.3 & 59.9 & 68.8 & 64.3 \\
\tbdots{genGray}
\rGen & \qwen\tbgap Qwen3-VL-32B & 75.4 & 79.7 & 70.9 & 42.3 & 75.3 \\
\tbdots{genGray}
\rGen & \qwen\tbgap Qwen3-VL-32B + Think & 67.8 & 75.5 & 68.5 & \underline{77.4} & 70.6 \\
\tbdots{genGray}
\rGen & \gemma\tbgap Gemma-4-31B & \underline{77.6} & \underline{81.5} & \underline{81.0} & 64.1 & \underline{80.0} \\
\tbsepc{genGray}
\rGen \hspace{0.1em} Kris~\cite{wu2025krisbench} & \qwen\tbgap Qwen3-VL-8B & 69.7 & 75.0 & 65.6 & 54.3 & 70.1 \\
\tbdots{genGray}
\rGen & \gemma\tbgap Gemma-4-31B & 75.3 & 78.9 & 79.0 & 64.3 & 77.7 \\
\tbsepc{genGray}
\rGen \hspace{0.1em} ImgEdit~\cite{ye2025imgedit} & \qwen\tbgap Qwen3-VL-8B & 70.2 & 72.9 & 69.3 & 44.3 & 70.8 \\
\tbdots{genGray}
\rGen & \gemma\tbgap Gemma-4-31B & 74.7 & 78.7 & 78.6 & 71.9 & 77.3 \\
\tbparadigmType{propGray}{\color{black!55}Generative scoring (proprietary models)}  \\
\rProp \hspace{0.1em} VIE-score & \faded{\gemini}\tbgap Gemini-2.5-flash & 77.0 & 82.8 & 82.1 & 63.1 & 80.6 \\
\tbdots{propGray}
\rProp & \faded{\openai}\tbgap GPT-4.1-mini & 73.7 & 78.1 & 70.5 & 49.9 & 74.1 \\
\tbdots{propGray}
\rProp & \faded{\openai}\tbgap GPT-4o & 78.3 & 80.9 & 74.6 & 47.6 & 77.9 \\
\tbdots{propGray}
\rProp & \faded{\openai}\tbgap GPT-5.4 & 84.1 & 87.0 & 83.2 & 63.5 & 84.8 \\
\tbdots{propGray}
\rProp & \faded{\openai}\tbgap GPT-5.6-sol & 88.7 & 92.6 & 87.3 & 86.6 & 89.5 \\
\tbdots{propGray}
\rProp & \faded{\claude}\tbgap Opus 5 & 87.8 & 89.8 & 90.4 & 73.2 & 89.3 \\
\tbparadigmType{probGray}{Probabilistic scoring} \\
\rProb \hspace{0.1em} Qwen-Reranker & \qwen\tbgap Qwen3-VL-Reranker-8B & 72.5 & 76.1 & 68.7 & 59.3 & 72.5 \\
\tbsepc{probGray}
\rProb \hspace{0.1em} Log-prob & \qwen\tbgap Qwen3-VL-4B & 72.6 & 70.9 & 69.3 & 53.4 & 70.9 \\
\tbdots{probGray}
\rProb & \qwen\tbgap Qwen3-VL-8B & 76.9 & 75.0 & 72.2 & 60.6 & 74.7 \\
\tbdots{probGray}
\rProb & \qwen\tbgap Qwen2.5-VL-7B & 76.0 & 74.6 & 69.3 & 63.1 & 73.3 \\
\tbdots{probGray}
\rProb & \gemma\tbgap Gemma-4-31B & \textbf{80.6} & \textbf{83.9} & \textbf{84.5} & \textbf{82.6} & \textbf{83.0} \\
\tbparadigmType{supGray}{Supervised scoring} \\
\rSup \hspace{0.1em} EditScore~\cite{luo2025editscore} & \qwen\tbgap Qwen3-VL-4B & 64.6 & 70.8 & 73.5 & 54.1 & 69.6 \\
\tbsepc{supGray}
\rSup \hspace{0.1em} ImgEdit-Judge~\cite{ye2025imgedit} & \qwen\tbgap Qwen2.5-VL-7B & 55.3 & 60.2 & 63.8 & 33.3 & 59.8 \\
\tbsepc{supGray}
\rSup \hspace{0.1em} EditReward~\cite{wu2025editreward} & \qwen\tbgap Qwen2.5-VL-7B & 70.6 & 72.9 & 78.0 & 54.7 & 73.8 \\
\bottomrule
\end{NiceTabular}
\end{adjustbox}
\end{table}

\subsection{Alignment with Human Preference Benchmarks}
\label{app:preference-alignment}
Because \benchmark relies on controlled, synthetic image-instruction pairs to isolate factual editing errors, a natural question is whether performance on our benchmark transfers to real-world evaluation. To test this, we evaluate a shared set of judge configurations across \benchmark and two real-image human preference datasets: GenAI-Bench~\cite{jiang2024genai} and EditRewardBench~\cite{wu2025editreward}. Rather than positioning \benchmark as a replacement for open-ended preference benchmarks, this analysis tests whether the failure modes surfaced by controlled factual audits also govern how well judges align with human preferences on real images.
\paragraph{Benchmarks and Metrics.}
While \benchmark uses counterfactual pairs to evaluate strict factual verification (measured via overall AUROC), GenAI-Bench and EditRewardBench capture pairwise human preferences between candidate edits generated from the same real source image and prompt. On these preference datasets, we compute pairwise accuracy: a judge is scored as correct if they assign a higher score to the human-preferred edit, receive half credit for ties, and human-tied pairs are excluded. For EditRewardBench, we evaluate agreement against the overall preference score. We report the metrics in Table~\ref{tab:bridge_matched_results}.

\paragraph{Rank Correlation.}
To compare judge performance across different metrics and scales, we compute Spearman's rank correlation ($\rho$) across all matched judge configurations (Table~\ref{tab:bridge_rank_correlation}). Despite using purely synthetic data, judge rankings on \benchmark correlate strongly with both real-image preference benchmarks ($\rho = 0.95$ and $\rho = 0.71$). This demonstrates that controlled factual verification effectively captures core capabilities required for real-world human preference alignment.

\providecommand{\rk}[1]{{\color{black!45}\scriptsize #1}}

\begin{table}[t]
\centering
\caption{\benchmark{} rankings agree with human-preference benchmarks.
Each judge uses the same template, image placement, and scoring on all three benchmarks, except where marked †.
For \benchmark{} we report Overall AUROC; for GenAI-Bench and EditRewardBench, pairwise human-agreement accuracy on pairs with a strict human preference (555 and 1,453 pairs).
\#: rank among the eleven judges (1 = best; ties share the average rank).
Spearman correlations between the rankings are given in Tab~\ref{tab:bridge_rank_correlation}.
\textbf{Bold}: best, \underline{underline}: second best.}
\label{tab:bridge_matched_results}
\footnotesize
\setlength{\tabcolsep}{3.5pt}
\setlength{\aboverulesep}{0pt}
\setlength{\belowrulesep}{0pt}
\begin{adjustbox}{max width=\linewidth}
\renewcommand{\tbdots}[1]{\tbdotcline{2-8}}
\NiceTab{l@{\hspace{6pt}}l c@{\hspace{4pt}}c @{\hspace{12pt}}c@{\hspace{4pt}}c @{\hspace{12pt}}c@{\hspace{4pt}}c}
\toprule
 & & \multicolumn{2}{c}{\textbf{\benchmark}} & \multicolumn{2}{c}{\textbf{GenAI-Bench}} & \multicolumn{2}{c}{\textbf{EditRewardBench}} \\
\cmidrule(lr){3-4}\cmidrule(lr){5-6}\cmidrule(lr){7-8}
\textbf{Method} & \textbf{Base Model} & AUROC & \rk{\#} & Acc. & \rk{\#} & Acc. & \rk{\#} \\
\midrule
\tbparadigmCorr{genGray}{Generative scoring} \\
\rGen \hspace{0.1em} VIE-score~\cite{ku2024viescore} & \llava\tbgap LLaVA-OV-7B & 51.9 & \rk{10} & 58.4 & \rk{11} & 48.9 & \rk{11} \\
\tbdots{genGray}
\rGen & \qwen\tbgap Qwen2.5-VL-7B & 55.8 & \rk{9} & 62.3 & \rk{10} & 50.7 & \rk{10} \\
\tbdots{genGray}
\rGen & \qwen\tbgap Qwen3-VL-4B & 51.5 & \rk{11} & 78.1 & \rk{9} & 55.8 & \rk{9} \\
\tbdots{genGray}
\rGen & \qwen\tbgap Qwen3-VL-8B & 56.6 & \rk{8} & 79.4 & \rk{8} & 56.1 & \rk{8} \\
\tbdots{genGray}
\rGen & \qwen\tbgap Qwen3-VL-32B & 65.3 & \rk{7} & 81.9 & \rk{6} & 59.1 & \rk{3} \\
\tbdots{genGray}
\rGen & \gemma\tbgap Gemma-4-31B & \underline{74.3} & \rk{2} & \textbf{83.3} & \rk{1.5} & \underline{59.5} & \rk{2} \\
\tbparadigmCorr{probGray}{Probabilistic scoring} \\
\rProb \hspace{0.1em} Qwen-Reranker & \qwen\tbgap Qwen3-VL-Reranker-8B & 67.1 & \rk{4} & \underline{83.1} & \rk{3} & 58.3 & \rk{4} \\
\tbsepc{probGray}
\rProb \hspace{0.1em} Log-prob & \qwen\tbgap Qwen2.5-VL-7B & 65.8 & \rk{6} & 81.8 & \rk{7} & 56.2 & \rk{7} \\
\tbdots{probGray}
\rProb & \qwen\tbgap Qwen3-VL-8B & 67.5 & \rk{3} & 82.2 & \rk{4.5} & 56.8 & \rk{6} \\
\tbdots{probGray}
\rProb & \gemma\tbgap Gemma-4-31B$^{\dagger}$ & \textbf{81.0} & \rk{1} & \textbf{83.3} & \rk{1.5} & 57.6 & \rk{5} \\
\tbparadigmCorr{supGray}{Supervised scoring} \\
\rSup \hspace{0.1em} EditReward~\cite{wu2025editreward} & \qwen\tbgap Qwen2.5-VL-7B & 66.3 & \rk{5} & 82.2 & \rk{4.5} & \textbf{60.6} & \rk{1} \\
\bottomrule
\end{NiceTabular}
\end{adjustbox}
\vspace{2pt}

\raggedright\scriptsize
Log-prob rows use the binary template that was run on all three benchmarks, so their \benchmark{} values differ from Tab.~\ref{tab:evaluation_results}.
$^{\dagger}$On \benchmark{}, the yes/no-question variant of the images-last prompt (as in Tab.~\ref{tab:evaluation_results}); on GenAI-Bench and EditRewardBench, the same prompt without the question.
\end{table}
\begin{table}[t]
\centering
\caption{\textbf{Rank agreement across matched judge configurations.}
We compute Spearman rank correlation over the 11 judge configurations shared across \benchmark, GenAI-Bench, and EditRewardBench.}
\label{tab:bridge_rank_correlation}
\footnotesize
\setlength{\tabcolsep}{6pt}
\begin{tabular}{@{}lc@{}}
\toprule
\textbf{Benchmark Pair} & \textbf{Spearman $\rho$} \\
\midrule
\benchmark{} vs. GenAI-Bench & 0.95 \\
\benchmark{} vs. EditRewardBench & 0.71 \\
GenAI-Bench vs. EditRewardBench & 0.83 \\
\bottomrule
\end{tabular}
\end{table}

\subsection{Human Validation - Multiple Choice}
\label{app:human}

Because ground-truth labels in \benchmark are determined programmatically from Blender rendering parameters, human annotation is not required to construct the dataset. Instead, we conduct a human study to verify that our synthetic edits and textual instructions are perceptually unambiguous, establishing a human performance ceiling for our multiple-choice diagnostic (Table~\ref{tab:evaluation_results}).

\begin{figure}
    \centering
    \includegraphics[width=.7\linewidth]{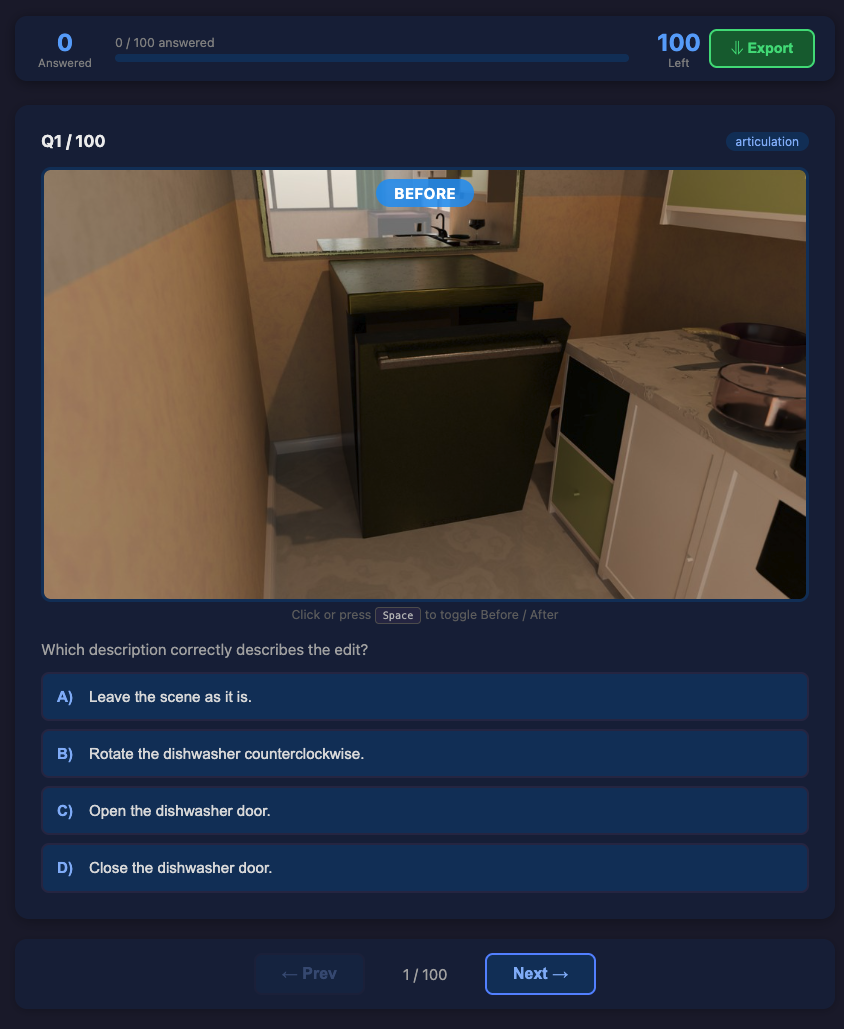}
    \caption{\textbf{Human validation interface.}
Participants toggled between the source and edited image and selected the instruction that correctly described the observed edit from candidate descriptions. The options included one factual instruction and multiple counterfactual distractors.}
    \label{fig:mc-interface}
\end{figure}

\paragraph{Study Design and Interface.}
Using the interface shown in Fig.~\ref{fig:mc-interface}, participants were presented with a source/edited image pair alongside one ground-truth instruction and a number of counterfactual distractors (spanning wrong-object, directional, cross-edit-type, and no-edit alternatives). To facilitate fine-grained visual comparison, participants could toggle back and forth between the \emph{Before} and \emph{After} views in place by clicking the image or pressing the spacebar before locking in the description that best matched the observed change.

\paragraph{Results.}
Five participants each evaluated 100 randomly sampled instances, achieving an average accuracy of 95.2\%. This near-ceiling performance confirms that the rendered edits are visually clear and unambiguously described. Moreover, it highlights a stark capability gap: while the task is easy for humans (95.2\%), the strongest VLM judge reaches only 71.5\% accuracy on the same forced-choice setup (Table~\ref{tab:evaluation_results}).

\paragraph{MC-style forced-choice diagnostic.}
In addition to AUROC, we evaluate a stricter within-example ranking diagnostic. For each before/after pair, we treat the factual instruction and its associated counterfactual instructions as a multiple-choice candidate set. Because several score-based judges produce ties, we report accuracy that gives $1/k$ credit when the factual instruction ties for first among $k$ top-scoring candidates.

Table~\ref{tab:evaluation_results} shows that this forced-choice diagnostic is substantially harder than AUROC. Even strong AUROC models often fail to select the factual instruction uniquely, especially when their scoring template produces coarse discrete scores: Kris with Gemma-4-31B reaches 72.9 AUROC but only 46.4\% MC accuracy, as its 1--5 scores often tie the factual instruction with distractors. AUROC thus captures aggregate positive--negative separability, while MC-style accuracy exposes whether a judge can make a clean per-example decision among matched distractors.

\subsection{Image Placement Ablation}
\label{app:image-placement}

Prior zero-shot edit-evaluation templates do not always specify the ordering of image and text tokens. In the official VIE-score implementation, we observed that Gemini prompts place the source and edited images before the textual instruction, whereas GPT prompts place the images after the text. We therefore evaluate both orderings while keeping the scoring template, image pairs, candidate instructions, and parsing code fixed.

Table~\ref{tab:image_placement_ablation} shows that this prompt-ordering choice has a large effect on measured judge performance. For Gemini-2.5-flash, moving the images after the text improves overall AUROC from 59.1 to 72.8. For GPT-4.1-mini, the same change improves AUROC from 53.3 to 67.4. The gains are broad across edit categories, especially for removal, material, movement, and rotation. These results indicate that multimodal prompt formatting is a consequential part of the evaluation protocol.

\begin{table}[t]
\centering
\caption{\textbf{The impact of image placement on zero-shot judge performance using VIE-score.}
Placing images after the instructional text substantially improves AUROC for both models.}
\label{tab:image_placement_ablation}
\setlength{\tabcolsep}{4pt}
\footnotesize
\begin{tabular}{@{}ll ccccccccc c@{}}
\toprule
\textbf{Model} & \textbf{Images} & \textbf{Rem} & \textbf{Mat} & \textbf{Lgt} & \textbf{Cam} & \textbf{Mov} & \textbf{Scl} & \textbf{Shp} & \textbf{Art} & \textbf{Rot} & \textbf{Overall} \\
\midrule
\multirow{2}{*}{Gemini-2.5-flash}
& Before Text & 69.0 & 67.9 & 67.9 & 49.6 & 57.8 & 54.8 & 56.9 & 59.7 & 48.0 & 59.1 \\
& After Text  & 94.6 & 86.2 & 83.8 & 71.8 & 76.5 & 57.1 & 58.2 & 72.8 & 54.0 & 72.8 \\
\midrule
\multirow{2}{*}{GPT-4.1-mini}
& Before Text & 57.4 & 62.5 & 70.6 & 53.8 & 59.2 & 42.8 & 41.3 & 53.8 & 37.9 & 53.3 \\
& After Text  & 80.1 & 81.1 & 79.8 & 59.2 & 71.3 & 57.8 & 54.9 & 67.2 & 55.0 & 67.4 \\
\bottomrule
\end{tabular}
\end{table}
\section{Dataset Generation Details}
\label{app:dataset-details}

\paragraph{Pipeline overview.}
We generate before/after image pairs by programmatically manipulating 3D indoor scenes produced by Infinigen~\cite{infinigen2024indoors}. Each scene is a \texttt{.blend} file containing a procedurally generated indoor environment with physically plausible object placement, materials, and lighting. Our pipeline has two decoupled phases. In the \emph{sampling} phase, we sample valid views and edit parameters using rejection sampling. In the \emph{rendering} phase, we render the resulting source and edited images with Blender's ray-tracing engine. This decoupling lets us validate edits in 3D before launching render jobs.

\paragraph{Scene representation.}
Each scene is parsed into a structured scene graph. Every object has a record containing its Blender mesh name, factory type, parent relation (floor, wall, or object), spatial relationships, and child objects. To increase object diversity, we deduplicate by object factory type when selecting edit targets, retaining at most one instance per factory class within a scene.

\paragraph{View selection.}
For each candidate target object, we attempt to place a virtual camera on a cylinder centred at the object's geometric centroid. The camera distance is sampled as $d \sim \mathcal{U}(d_{\min}, d_{\max})$, the azimuth is sampled within $\pm45^\circ$ of the object's forward-facing direction, and the height is sampled as $z \sim \mathcal{U}(z_{\min}, z_{\max})$. The camera is constrained to look at the target object. A view is accepted only if the target visibility exceeds $\tau_{\mathrm{vis}}=0.2$, estimated by face-level ray casting with frustum culling, backface culling, and occlusion checks.

Within each accepted view, we identify visible objects using both a face-sampling visibility test and a Monte Carlo pixel-area test. Objects must pass a visibility threshold of at least $0.3$ and occupy at least $0.04$ of the sampled image area. We then sample edits over all enabled edit types, resetting the scene to its original state between edits so that each before/after pair is independent.

\paragraph{Physical validation.}
Spatial and shape-changing edits are validated with rejection sampling. We use collision checks to reject edits that cause interpenetration, visibility checks to ensure that the edited object remains visible, and minimum-magnitude thresholds to ensure perceptibility. Collision tests use cached object geometry and bounding volumes to avoid repeatedly rebuilding the full scene during sampling.

\subsection{Edit Types}
\label{app:edit-types}

\benchmark contains nine benchmark edit categories: scale, removal/addition, material, movement, shape, rotation, articulation, lighting, and camera. Camera edits contain two subtypes, zoom and view, which are both reported under the camera category in the main experiments.

\textbf{Scale.} We apply a uniform scale factor $s \sim \mathcal{U}(0.5, 2.0)$ to all axes. Edits that cause collisions or leave the object insufficiently visible are rejected. Wall-mounted objects are excluded.

\paragraph{Removal and addition.}
We hide a prominently visible object with visibility at least $0.5$ to produce removal edits. Addition edits are obtained by swapping the before and after images of a removal edit; all benchmark pairs use the removal direction.

\paragraph{Material.}
We sample a physically based rendering material from a curated texture library and apply it to the target object. Each material may include diffuse, normal, roughness, metallic, and displacement maps. For objects with semantically distinct material regions, such as appliance bodies, handles, or glass panels, we also support per-slot material replacement by modifying the object's factory parameter dictionary and rebuilding the mesh.

\paragraph{Movement.}
For a visible object with parent surface $S$, we sample $N=200$ candidate positions on $S$ by surface point sampling. A candidate is accepted only if the displacement is at least $0.3$m, the object does not collide with other scene geometry, and the object remains visible from the camera. Wall-mounted objects are reoriented to align with the new surface normal. Movement direction is computed relative to the camera view plane and quantised into eight compass directions.

\paragraph{Shape.}
Shape edits modify continuous procedural factory parameters, such as object width, height, depth, or chair back height. A new parameter value is sampled within an edit range extending beyond the default factory sampling range, and the relative change must exceed $5\%$ of the edit range. The edited object is rebuilt at its original world-space transform and checked for collisions. If the edited object has children, they are repositioned to maintain physical support when possible; children that still collide after repositioning are hidden.

\paragraph{Rotation.}
We rotate the target object about the vertical axis by $\alpha \sim \mathcal{U}(-160^\circ,160^\circ)$ with $|\alpha|\geq10^\circ$. Rotations that cause collisions are rejected, with up to 50 resampling attempts. Wall-mounted objects are excluded.

\paragraph{Articulation.}
Articulation edits modify joint parameters, such as oven or fridge door rotation. The object is rebuilt from its factory with the modified joint value. Since articulation may change the bounding box while the object body should remain fixed, we compute an origin correction from a reference mesh at the original parameter value and apply the corresponding offset to keep the body stationary.

\paragraph{Camera.}
Camera edits include two subtypes. \emph{Zoom} perturbs the focal length by $\Delta f \sim \mathcal{U}(\pm2,\pm6)$mm. \emph{View} re-places the camera on the viewing cylinder around the same target object, requiring a camera-position displacement between $0.3$m and $1.0$m.

\paragraph{Lighting.}
Lighting edits apply one of 10 named illumination presets, such as \texttt{warm}, \texttt{sunset}, or \texttt{night}. Each preset specifies RGB color and energy values that are applied to the point lights in the scene.

\section{Taxonomy Construction Methodology}
\label{app:taxonomy-method}

This appendix details the data-driven pipeline used to construct the
editing operation taxonomy summarised in
Section~\ref{sec:taxonomy}.

\subsection{Data Source}
\label{app:taxonomy-data}

We base our analysis on the RealEdit dataset~\cite{sushko2025realedit}, which
aggregates image editing requests posted by real users on the Reddit
community \texttt{r/PhotoshopRequest}. Unlike synthetically
constructed benchmarks, these instructions capture the full
diversity, ambiguity, and complexity of how non-expert users
communicate editing intent. After filtering the training split to
retain only posts with non-empty instruction fields, we obtain a
corpus of approximately 42{,}400 unique editing instructions.

\subsection{Embedding and Diversity-Aware Sampling}
\label{app:taxonomy-embedding}

We encode all instructions using
Qwen3-Embedding-8B~\cite{zhang2025qwen3}, an 8-billion-parameter
instruction-tuned text embedding model, prompted with the task
description \emph{``Given an image editing instruction, retrieve
instructions that belong to the same editing category or require
similar photographic modifications.''}

To obtain a representative yet diverse subset for manual inspection,
we apply greedy farthest-point sampling in cosine distance space:
\begin{equation}
    x_{k+1} = \arg\max_{x \in \mathcal{X} \setminus \mathcal{S}_k}
    \min_{s \in \mathcal{S}_k} d_{\cos}(x, s),
    \label{eq:fps}
\end{equation}
where $\mathcal{S}_k$ is the set of $k$ selected instructions and
$d_{\cos}$ is cosine distance. This yields 1{,}024 instructions
that uniformly cover the embedding space.

\subsection{Cluster Analysis}
\label{app:taxonomy-clustering}

We apply Bisecting $k$-Means~\cite{scikit-learn} at multiple granularities
($k \in \{32, 64, 128, 256, 512, 1024, 2048\}$) on the full
42{,}400 instruction embeddings. Within each cluster, we subsample
up to 512 points and apply farthest-point selection
(Eq.~\ref{eq:fps}) to extract 16 maximally diverse
representatives for qualitative inspection. At $k{=}64$, coherent
clusters emerged for animal-related edits, text and logo
manipulation, person-centric edits, and scene/environment
modification, among others.

\subsection{LLM-Based Instruction Decomposition}
\label{app:taxonomy-decomp}

Clustering reveals \emph{what kinds} of instructions co-occur, but
does not identify the atomic operations within each instruction. We
prompt a 120-billion-parameter open-source LLM to decompose each
instruction into a JSON object listing one or more atomic
operations, each characterised by:
\begin{itemize}[nosep]
    \item \textbf{action}: a normalised verb (e.g.,
    \texttt{remove}, \texttt{change\_color}, \texttt{add}),
    \item \textbf{target}: the entity being edited,
    \item \textbf{target\_type}: one of \{\texttt{object},
    \texttt{object\_part}, \texttt{region}, \texttt{attribute},
    \texttt{quality\_defect}, \texttt{style}, \texttt{global}\},
    \item \textbf{parameters}: additional modifiers.
\end{itemize}
The decomposition prompt was refined through three iterations of
prompt engineering and manual validation against cluster exemplars.

\subsection{Embedding-Based Retrieval Validation}
\label{app:taxonomy-retrieval}

For each candidate operation category, we construct 5--18
representative query instructions and retrieve the top-100 most
similar instructions from the full corpus by cosine similarity.
Manual inspection of retrieved results served two purposes:
(1)~coverage validation and (2)~boundary refinement between
categories.

\subsection{The Minimalist Editing Language (MEL)}
\label{app:mel}

The analyses above converge on a structured editing representation
language. MEL defines three primitive operations:
\begin{itemize}[nosep]
    \item \texttt{SET(attribute, target, description)} --- modifies
    an abstract quality of an existing entity.
    \item \texttt{ADD(content, location)} --- creates a new element.
    \item \texttt{REMOVE(target)} --- deletes a bounded element.
\end{itemize}
Two convenience macros express common composites:
\begin{itemize}[nosep]
    \item \texttt{REPLACE(target, new\_content)} $\rightarrow$
    \texttt{REMOVE(target)}; \texttt{ADD(new\_content,
    previous\_position)}.
    \item \texttt{SWAP(obj1, obj2)} $\rightarrow$
    \texttt{SET(position, obj1, obj2's position)};
    \texttt{SET(position, obj2, obj1's position)}.
\end{itemize}

\paragraph{Entity types.} Each command specifies an entity type:
\textbf{Object} (a discrete bounded entity), \textbf{Environment}
(ambient scene-wide properties), or \textbf{Image} (global
post-processing properties).

\paragraph{The ``Can You Circle It?'' rule.} The most frequent
disambiguation challenge is distinguishing removal from
modification: \emph{``remove the blur''} does not mean deleting a
bounded object but adjusting the sharpness attribute. MEL resolves
this with a simple heuristic: if the target has a drawable boundary,
use \texttt{REMOVE}; if it is an abstract quality, use \texttt{SET}.

\subsection{Attribute Taxonomy}
\label{app:attributes}

A large fraction of editing instructions involve modifying
\emph{attributes} of existing entities. The LLM decompositions
produced over 1{,}000 unique attribute terms. We group them into 20
canonical categories through frequency analysis, semantic
similarity, and expert judgement:

\begin{enumerate}[nosep]
    \small
    \item Color
    \item Light \& Exposure
    \item Focus \& Clarity
    \item Texture \& Surface
    \item Grain \& Noise
    \item Opacity \& Transparency
    \item Digital Artifacts
    \item Position \& Alignment
    \item Size \& Dimensions
    \item Framing \& Cropping
    \item Scene \& Environment
    \item Identity \& Appearance
    \item Pose \& Expression
    \item Anatomy \& Features
    \item Clothing \& Attire
    \item Text \& Typography
    \item Blemish \& Flaw Removal
    \item Aesthetic \& Style
    \item Overall Quality \& Appeal
    \item Effects \& Filters
\end{enumerate}

This mapping covers over 280 distinct surface-level terms observed
in user instructions.

\subsection{Large-Scale Corpus Annotation}
\label{app:annotation}

Using the finalised MEL specification, we translate all 42{,}400
instructions into structured MEL representations via batched
inference (batch size 1{,}024; temperature $T{=}0.7$; max tokens
1{,}024) with the MEL specification as the system prompt.
\section{Prompt Templates}
\label{app:prompts}

This appendix reproduces the full prompt text for each of the
templates compared in Section~\ref{sec:experiments}.

\subsection{Binary Probability Baseline}
\label{app:prompt-binary}

\begin{promptbox}{System prompt}
\small\ttfamily
You are an expert at evaluating image edits. Answer with only
\texttt{True} or \texttt{False}.
\end{promptbox}

\begin{promptbox}{User prompt}
\small\ttfamily
Edit instruction: ``\{instruction\}''

\vspace{4pt}
\{question\}

\vspace{4pt}
Before Image:

\textnormal{\textit{[before image]}}

After Image:

\textnormal{\textit{[after image]}}

\vspace{4pt}
Answer with only \texttt{True} or \texttt{False}.
\end{promptbox}

\texttt{\{question\}} is the instruction rephrased as a yes/no question (e.g., ``Move the cell shelf right.'' $\rightarrow$ ``Has the cell shelf been moved right?''); the no-question variant omits this line.

\paragraph{Parsing.}
For sampled responses, we scan the output case-insensitively for the tokens
\texttt{true} and \texttt{false}. If both appear, the last occurrence determines
the prediction. For probability scoring, we use the next-token probability
assigned to \texttt{True}.

\subsection{VIEScore Prompt}
\label{app:prompt-viescore}

\begin{promptbox}{System prompt}
\small\ttfamily
You are an expert evaluator of image edits.
\end{promptbox}

\begin{promptbox}{User prompt}
\small\ttfamily
Editing instruction: \{instruction\}

\vspace{4pt}
\textnormal{\textit{[before image]}}
\textnormal{\textit{[after image]}}

\vspace{4pt}
From scale 0 to 10: A score from 0 to 10 will be given based on
the success of the editing. (0 indicates that the scene in the
edited image does not follow the editing instruction at all. 10
indicates that the scene in the edited image follows the editing
instruction text perfectly.) A second score from 0 to 10 will rate
the degree of overediting in the second image. (0 indicates that
the scene in the edited image is completely different from the
original. 10 indicates that the edited image can be recognized as a
minimally edited yet effective version of the original.) Put the
score in a list such that output score = [score1, score2], where
\texttt{score1} evaluates the editing success and \texttt{score2}
evaluates the degree of overediting.
\end{promptbox}

\paragraph{Parsing.}
We extract the first two-element numerical list from the response using a
regular expression, e.g.\ \texttt{[8, 7]}. The continuous score used for AUROC
is the arithmetic mean of the two values.

\subsection{KRIS Instruction-Following Prompt}
\label{app:prompt-kris}

\begin{promptbox}{User prompt}
\scriptsize\ttfamily
You are a professional digital artist and image evaluation specialist. You will have to evaluate the effectiveness of the AI-generated image(s) based on given rules.

You will be given:

1. \textbf{Image A}: the original image.

2. \textbf{Image B}: an edited version of Image A.

3. \textbf{Editing Instruction}: a directive describing the intended modification to Image A to produce Image B.

Your Objective:

Your task is to \textbf{evaluate how the edited image faithfully fulfills the editing instruction}, focusing \textbf{exclusively on the presence and correctness of the specified changes}.

You must:

\textbf{Identify detailed visual differences} between Image A and Image B \textbf{correctly and faithfully}.

Determine if those differences \textbf{match exactly what the editing instruction requests}.

\textbf{Not assess any unintended modifications beyond the instruction}; such evaluations fall under separate criteria, e.g., visual consistency.

\textbf{Be careful}, an edit may introduce visual change without fulfilling the actual instruction, e.g., replacing the object instead of modifying it.

\vspace{4pt}
\textbf{\#\# Reasoning:}

You must follow these reasoning steps before scoring:

\textbf{1. Detect Difference}: What has visually changed between Image A and Image B? In this step, you do not have to use information from the editing instruction.

\textbf{2. Expected Visual Caption}: Write a factual description of how the edited image should look if the instruction were perfectly followed.

\textbf{3. Instruction Match}: Compare the observed differences in 1 to the expected change in 2:
Was the correct object modified, not replaced? Was the requested attribute modified as intended? Is the degree of modification accurate?

\textbf{4. Decision}: Use the 1--5 scale to assign a final score.

\vspace{4pt}
\textbf{\#\# Evaluation Scale (1 to 5):}

\textbf{5 Perfect Compliance}: The edited image precisely matches the intended modification.

\textbf{4 Minor Omission}: The core change is made, but a minor detail is missing or slightly incorrect.

\textbf{3 Partial Compliance}: The main idea is present, but one or more required aspects are wrong or incomplete.

\textbf{2 Major Omission}: Most required changes are missing or poorly implemented.

\textbf{1 Non-Compliance}: The instruction is not followed at all or is completely misinterpreted.

\vspace{4pt}
\textbf{\#\# Input}

\textbf{Image A}

\textbf{Image B}

\textbf{Editing Instruction}: \{instruction\}

\vspace{4pt}
\textbf{\#\# Output Format}

Look at the input again, provide the evaluation score and explanation in JSON format:

\{
``instruction\_score'': X,

``reasoning'': ``1. Detect Difference 2. Expected Visual Caption 3. Instruction Match 4. Decision''
\}

\vspace{4pt}
This is the original image:

\textnormal{\textit{[before image]}}

This is the edited image:

\textnormal{\textit{[after image]}}
\end{promptbox}

\paragraph{Parsing.}
We parse the JSON field \texttt{instruction\_score}. If JSON parsing fails, we fall back to a case-insensitive scan for \texttt{instruction score}. The resulting 1--5 score is divided by 5 to obtain the continuous judge score used for AUROC.

\subsection{ImgEdit Prompt}
\label{app:prompt-imgedit}

\begin{promptbox}{System prompt - ImgEdit}
\small\ttfamily
You are an expert evaluator of image edits.
\end{promptbox}

\begin{promptbox}{User prompt - ImgEdit}
\small\ttfamily
Instruction: ``\{instruction\}''

\vspace{4pt}
Source Image:

\textnormal{\textit{[before image]}}

Edited Image:

\textnormal{\textit{[after image]}}

\vspace{4pt}
Please evaluate the edited image based on the following criteria:

\vspace{4pt}
1. Prompt Compliance (1--5): Does the edit follow the instruction?

\hspace*{1em}5: Perfectly follows instruction

\hspace*{1em}4: Mostly follows instruction with minor deviations

\hspace*{1em}3: Partially follows instruction

\hspace*{1em}2: Barely follows instruction

\hspace*{1em}1: Does not follow instruction

\vspace{4pt}
2. Visual Seamlessness (1--5): How natural and seamless is the edit?

\hspace*{1em}5: Perfectly seamless, no visible artifacts

\hspace*{1em}4: Mostly seamless with minor imperfections

\hspace*{1em}3: Noticeable but acceptable seams

\hspace*{1em}2: Obvious visual artifacts

\hspace*{1em}1: Very poor visual quality

\vspace{4pt}
3. Physical \& Detail Fidelity (1--5): Are physical properties and
details preserved?

\hspace*{1em}5: Perfect physical realism and detail preservation

\hspace*{1em}4: Good realism with minor detail loss

\hspace*{1em}3: Acceptable but some unrealistic elements

\hspace*{1em}2: Noticeable physical inconsistencies

\hspace*{1em}1: Severely unrealistic

\vspace{4pt}
Format your response as:

\texttt{Prompt Compliance: [score]}

\texttt{Visual Seamlessness: [score]}

\texttt{Physical \& Detail Fidelity: [score]}
\end{promptbox}

\paragraph{Parsing.}
We extract each named dimension using regular expressions, e.g.\
\texttt{Prompt Compliance: 4}. The continuous score used for AUROC
is the arithmetic mean of all successfully parsed dimensions.

\section{Limitations}
\label{app:limitations}

\paragraph{Synthetic-to-real transfer.}
\benchmark uses synthetic indoor scenes rendered from procedurally generated Blender assets. This design provides exact ground truth over what changed, which is essential for auditing judge failures, but it also limits visual diversity relative to real image-editing benchmarks. Results on \benchmark should therefore be interpreted as diagnostic evidence about controlled edit verification, not as a complete substitute for evaluation on natural images.

\paragraph{Scope of edit types.}
The benchmark covers nine edit types: removal/addition, material, lighting, camera, movement, scale, shape, articulation, and rotation. These edits were chosen to cover common low-level and object-level transformations with controllable ground truth. However, they do not exhaust the space of real editing requests, especially complex semantic edits, identity-preserving edits, style transfer, text rendering, human-centric edits, and open-ended compositional changes.

\paragraph{Judge evaluation, not editor evaluation.}
Our benchmark audits VLM-based judges by asking whether they assign higher scores to factual edit instructions than to counterfactual ones. It does not directly measure the quality of image-editing models, since the edited images are generated by a controlled synthetic pipeline rather than by arbitrary editing systems. The findings therefore speak to the reliability of automated evaluation protocols, not to the absolute quality of any image editor.

\paragraph{Prompt and implementation sensitivity.}
We find that small implementation choices, such as image-text ordering, can substantially affect judge performance. Consequently, reported numbers should be understood as properties of a model, prompt template, scoring rule, and preprocessing stack together. Future model releases or different inference settings may change absolute performance, although the diagnostic structure of the benchmark remains applicable.

\paragraph{Controlled counterfactuals.}
Counterfactual instructions are generated from templates and parameter perturbations. This makes failures interpretable, but it may not cover all linguistic variability found in natural user instructions. In particular, real instructions may be ambiguous, underspecified, conversational, or contain multiple simultaneous edit requests.

\paragraph{Model coverage.}
We evaluate a broad set of recent VLM judges and scoring paradigms, including generative prompts, probabilistic scoring, reranking, and supervised judge models. However, the VLM landscape changes quickly, and proprietary model versions may change over time. Our results should be viewed as an audit of the evaluated systems and protocols rather than a permanent ranking of all possible judges.

\paragraph{Pointwise vs.\ pairwise evaluation.}
Recent benchmarks use pairwise VLM comparisons over pointwise scoring~\cite{li2026genarena,jiang2026geditbench}. We audit pointwise scoring to match our evaluated judges and benchmarks (VIEScore, EditScore, EditReward, ImgEdit-Judge, GEdit-Bench), leaving it open whether pairwise gains hold against ground truth rather than human preference. As a proxy, \benchmark{}'s MC accuracy (Tab.~\ref{tab:evaluation_results}) uses pointwise scores to rank factual instructions against matched counterfactuals, where the best open (60.8\%) and overall (71.5\%) judges trail humans (95.2\%). Extending our evaluation to image-side pairwise evaluation is left to future work.
\section{Benchmark Card}
\label{app:benchmark-card}

\paragraph{Motivation.}
\benchmark is designed to audit VLM-based image-editing judges under controlled ground truth. The benchmark targets factual edit verification: given a source image, an edited image, and an instruction, a judge should assign higher scores to factual instructions than to counterfactual instructions. It is not intended to evaluate the perceptual quality or user preference of image-editing models in the wild.

\paragraph{Composition.}
\benchmark contains 1,382 synthetic before/after edit pairs rendered in Blender across 1,000 procedurally generated indoor scenes. Edits span nine edit types: removal/addition, material, lighting, camera, movement, scale, shape, articulation, and rotation. Each edit pair is associated with factual and counterfactual text instructions, including directional, wrong-object, cross-edit-type, and no-edit counterfactuals.

\paragraph{Ground Truth.}
Because each scene is generated procedurally, the benchmark stores the exact Blender parameters before and after each edit. These include object position $\mathbf{x}$, rotation $\mathbf{r}$, scale $\mathbf{s}$, procedural shape parameter $\theta$, articulation parameter $\psi$, lighting state $\ell$, object visibility $v$, material assignment $m$, and camera state $\mathcal{C}$. These parameters enable edit-type-level analysis and salience-based stratification of judge failures.

\paragraph{Generation Process.}
Scenes are sampled from a procedural 3D pipeline and rendered with Blender. For each target object, valid views and edits are sampled using rejection criteria that enforce visibility, collision-free geometry, physical plausibility, and minimum perceptibility. Source and edited images are rendered from the resulting scene states. Factual instructions are generated from the known edit parameters, and counterfactual instructions are generated by perturbing the direction, target object, edit type, or no-edit hypothesis.

\paragraph{Human Validation.}
Although the structural ground truth is guaranteed by construction, we validate perceptual clarity with a multiple-choice human study. Participants toggle between source and edited images and select the matching factual instruction. On 100 representative edit triplets, five participants achieved 100\%, 98\%, 97\%, 92\%, and 89\% accuracy, suggesting that the rendered edits are perceptually unambiguous to humans.

\paragraph{Intended Uses.}
\benchmark is intended for evaluating and auditing VLM-as-a-judge systems, prompt templates, scoring rules, and edit-verification metrics. It is especially suited for diagnosing systematic judge failures by edit type, counterfactual type, and controlled edit magnitude.

\paragraph{Out-of-Scope Uses.}
The benchmark should not be interpreted as a complete measure of real-world image-editing quality, aesthetic preference, or generative model usefulness. Since all images are synthetic indoor scenes, results may not fully transfer to natural photographs, artistic images, human faces, or open-ended real user edits.

\paragraph{Limitations.}
The benchmark trades real-image diversity for controlled ground truth. It covers a fixed set of edit types and procedural scene distributions, and may underrepresent semantic, stylistic, or object categories common in real editing tasks. VLM behavior may also depend on prompt formatting, model version, and image preprocessing choices.

\paragraph{Ethical Considerations.}
The benchmark is synthetically generated and does not contain real personal data. However, automated judge scores may be misused as proxies for human preference or image quality. We therefore recommend using \benchmark as a diagnostic audit tool rather than as a standalone replacement for human evaluation.

\paragraph{Access and Maintenance.}
The dataset is available at \url{https://huggingface.co/datasets/JensParslov/EditJudge-Bench} (CC BY-NC 4.0) and the code at \url{https://github.com/JensPars/EditJudge-Bench} (MIT). The release includes the rendered images, instruction triplets, saved ground-truth parameters, evaluation scripts, and Croissant metadata. We version updates with git tags and document schema changes.

\subsection{Compute Resources}
\label{app:compute}

\paragraph{Dataset generation and rendering.}
Dataset generation consists of CPU-side edit sampling followed by Blender rendering. Edit sampling parses each Infinigen scene, samples candidate edits, and applies rejection sampling for visibility, collision, and validity constraints. Rendering was performed with Blender's Cycles renderer using GPU workers, primarily NVIDIA L40S GPUs, at $1920 \times 1080$ resolution. Rendering is embarrassingly parallel across scenes and edit candidates. The released benchmark contains all 1,500 rendered edit pairs (1,382 curated) as lossless $1920\times1080$ PNG renders.

\paragraph{Open-weight judge inference.}
Open-weight VLM judges were served with vLLM on a single NVIDIA H100 GPU per run. Where official quantized checkpoints were available, we used the corresponding quantized variants, listed in our released code. We enabled vLLM prefix caching, which is useful in our setting because candidate instructions for the same before/after pair share the same image inputs and much of the prompt prefix. This substantially reduces repeated prefill computation across the positive and counterfactual instructions for each image pair.

\paragraph{Hosted models and metric computation.}
Proprietary models were evaluated through hosted APIs rather than local GPU inference. Their runtime depends on provider-side batching, rate limits, and response length. After prediction files are produced, all metric computation is lightweight: AUROC aggregation, per-counterfactual-type breakdowns, salience stratification, MC-style accuracy, and bridge-analysis statistics run on CPU in minutes and require only a few GB of memory. The released evaluation code reproduces these aggregate metrics from saved prediction files without rerunning VLM inference or Blender rendering.

\end{document}